\def\isarxiv{1} %%%ICML version, we comment this line

\ifdefined\isarxiv
\documentclass[11pt]{article}
\else
\documentclass{article}
\usepackage{hyperref}
\usepackage{icml2021}
\fi

\usepackage{microtype}
\usepackage{amsmath}
\usepackage{amsthm}
\allowdisplaybreaks
\usepackage{amssymb}
\usepackage{algorithm}
\usepackage{subfig}
\usepackage{color}
\usepackage{epstopdf}
\usepackage{url}
\usepackage{graphicx}
\usepackage{color}
\usepackage{epstopdf}
\usepackage{algpseudocode}
\usepackage{scrextend}
\usepackage[T1]{fontenc}
\usepackage{bbm}
\usepackage{comment}

\usepackage{multicol}
\usepackage{multirow}
\usepackage{dsfont}
\usepackage{mathtools}
\usepackage{enumitem}

\usepackage[makeroom]{cancel}
\usepackage{stmaryrd}
\usepackage{booktabs, makecell}
\usepackage{pifont}% http://ctan.org/pkg/pifont
\usepackage{lipsum}

\usepackage{tikz}

\usetikzlibrary{arrows}
\ifdefined\isarxiv
\usepackage{hyperref}
\hypersetup{colorlinks=true,citecolor=red,linkcolor=red}
\usepackage[margin=1in]{geometry}
\else

\fi
\graphicspath{{./figs/}}

\ifdefined\isarxiv
\usepackage[margin=1in]{geometry}
\else

\fi

\definecolor{b2}{RGB}{51,153,255}
\definecolor{mygreen}{RGB}{80,180,0}
\definecolor{yl}{RGB}{255,80,0}
\definecolor{myl}{RGB}{180,80,20}

\newtheorem{theorem}{Theorem}[section]
\newtheorem{lemma}[theorem]{Lemma}
\newtheorem{definition}[theorem]{Definition}

\newtheorem{assumption}[theorem]{Assumption}

\newtheorem{claim}[theorem]{Claim}

\newcommand{\wt}{\widetilde}

\newcommand{\ov}{\overline}

\DeclareMathOperator{\poly}{poly}

\DeclareMathOperator{\R}{{\mathbb R}}

\DeclareMathOperator*{\E}{{\mathbb{E}}}

\newcommand{\loc}{\mathrm{local}}
\newcommand{\glo}{\mathrm{global}}
\DeclareMathOperator{\vect}{vec}

\newcommand{\N}{\mathcal{N}}

\ifdefined\isarxiv
\title{FL-NTK: A Neural Tangent Kernel-based Framework \\for Federated Learning Convergence Analysis\thanks{A preliminary version of this paper appeared in the Proceedings of the 38th International Conference on Machine Learning (ICML 2021).}}
\date{}

\author{
Baihe Huang\thanks{\texttt{baihehuang@pku.edu.cn}. Peking University.}
\and 
Xiaoxiao Li\thanks{\texttt{xl32@princeton.edu}. Princeton University.}
\and 
Zhao Song\thanks{\texttt{zhaos@princeton.edu}. Princeton University.}
\and
Xin Yang\thanks{\texttt{yx1992@cs.washington.edu}. The University of Washington.}
}

\else 
\icmltitlerunning{FL-NTK: A Neural Tangent Kernel-based Framework for Federated Learning Convergence Analysis}
\fi

\begin{document}

\ifdefined\isarxiv
\else
\twocolumn[
\icmltitle{FL-NTK: A Neural Tangent Kernel-based Framework \\for Federated Learning Analysis}



\icmlsetsymbol{equal}{*}

\begin{icmlauthorlist}
\icmlauthor{Aeiau Zzzz}{equal,to}
\icmlauthor{Bauiu C.~Yyyy}{equal,to,goo}
\icmlauthor{Cieua Vvvvv}{goo}
\icmlauthor{Iaesut Saoeu}{ed}
\end{icmlauthorlist}

\icmlaffiliation{to}{Department of Computation, University of Torontoland, Torontoland, Canada}
\icmlaffiliation{goo}{Googol ShallowMind, New London, Michigan, USA}
\icmlaffiliation{ed}{School of Computation, University of Edenborrow, Edenborrow, United Kingdom}

\icmlcorrespondingauthor{Cieua Vvvvv}{c.vvvvv@googol.com}
\icmlcorrespondingauthor{Eee Pppp}{ep@eden.co.uk}

\icmlkeywords{Machine Learning, ICML}

\vskip 0.3in
]



\printAffiliationsAndNotice{\icmlEqualContribution} % otherwise use the standard text.
\fi

\ifdefined\isarxiv

\begin{titlepage}
\maketitle
\begin{abstract}
Federated Learning (FL) is an emerging learning scheme that allows different distributed clients to train deep neural networks together without data sharing. Neural networks have become popular due to their unprecedented success. To the best of our knowledge, the theoretical guarantees of FL concerning neural networks with explicit forms and multi-step updates are unexplored. 
Nevertheless, training analysis of neural networks in FL is non-trivial for two reasons: first, the objective loss function we are optimizing is non-smooth and non-convex, and second, we are even not updating in the gradient direction.
Existing convergence results for gradient descent-based methods heavily rely on the fact that the gradient direction is used for updating.
This paper presents a new class of convergence analysis for FL, Federated Learning Neural Tangent Kernel (FL-NTK), which corresponds to overparamterized ReLU neural networks trained by gradient descent in FL
and is inspired by the analysis in Neural Tangent Kernel (NTK). 
Theoretically, FL-NTK converges to a global-optimal solution at a linear rate with properly tuned learning parameters. 
Furthermore, with proper distributional assumptions, FL-NTK can also achieve good generalization.

%The proposed theoretical analysis scheme can be generalized to more complex neural networks.

\end{abstract}
\thispagestyle{empty}
\end{titlepage}

\tableofcontents
\newpage

\else

\begin{abstract}
Federated Learning (FL) is an emerging learning scheme that allows different distributed clients to train deep neural networks together without data sharing. Neural networks have become popular due to their unprecedented success. To the best of our knowledge, the theoretical guarantees of FL concerning neural networks with explicit forms and multi-step updates are unexplored. 
Nevertheless, training analysis of neural networks in FL is non-trivial for two reasons: first, the objective loss function we are optimizing is non-smooth and non-convex, and second, we are even not updating in the gradient direction.
Existing convergence results for gradient descent-based methods heavily rely on the fact that the gradient direction is used for updating.
This paper presents a new class of convergence analysis for FL, Federated Learning Neural Tangent Kernel (FL-NTK), which corresponds to overparamterized ReLU neural networks trained by gradient descent in FL
and is inspired by the analysis in Neural Tangent Kernel (NTK). 
Theoretically, FL-NTK converges to a global-optimal solution at a linear rate with properly tuned learning parameters. 
Furthermore, with proper distributional assumptions, FL-NTK can also achieve good generalization.

%The proposed theoretical analysis scheme can be generalized to more complex neural networks.

\end{abstract}

\fi
\section{Introduction}
In traditional centralized training, deep learning models learn from the data, and data are collected in a database on the centralized server. In many fields, such as healthcare and natural language processing, models are typically learned from personal data. These personal data are subject to regulations such as California Consumer Privacy Act (CCPA) \cite{ccpa}, Health Insurance Portability and Accountability Act (HIPAA) \cite{act1996health}, and General Data Protection Regulation (GDPR) of European Union. Due to the data regulations, standard centralized learning techniques are not appropriate, and users are much less likely to share data. Thus, the data are only available on the local data owners (i.e. edge devices). Federated learning (FL) is a new type of learning scheme that avoids centralizing data in model training. FL allows local data owners (also known as clients) to locally train the private model and then send the model weights or gradients to the central server. Then central server aggregates the shared model parameters to update new global model, and broadcasts the the parameters of global model to each local client.

Quite different from the centralized training, FL has the following unique properties. First, the training data are distributed on an astonishing number of devices, and the connection between the central server and the device is slow. Thus, the computational cost is a key factor in FL. In communication-efficient FL, local clients are required to update model parameters for a few steps locally then send their parameters to the server \cite{mmr+17}. Second, due to the fact that the data are collected from different clients, the local data points can be sampled from different local distributions. When this happens during training, convergence may not be guaranteed. 

The above two unique properties not only bring challenges to algorithm design but also make theoretical analysis much harder. There have been many efforts developing convergence guarantees for FL algorithms based on the
assumptions of convexity and smoothness for the objective functions~\cite{yyz19,lhy+19,kmr20}. Although a recent study~\cite{ljz+21} shows theoretical studies of FL on neural networks, its framework fails to generate multiple-local updates analysis, which is a key feature in FL. One of the most conspicuous questions to ask is: 
\begin{center}
\textit{Can we build a unified and generalizable convergence analysis framework for ReLU neural networks in FL?}
\end{center}
In this paper, we give an affirmative answer to this question. It was recently proved in a series of papers that gradient descent converges to global optimum if the neural networks are overparameterized (significantly larger than the size of the datasets). Our work is motivated by Neural Tangent Kernel (NTK) that is originally proposed by \cite{jgh18} and has been extensively studied over the last few years.  

NTK is defined as the inner product space of pairwise data point gradient (aka Gram matrix). It describes the evolution of deep artificial neural networks during their training by gradient descent. Thus, we propose a novel NTK-based framework for federated learning convergence and generalization analysis on ReLU neural networks, so-called Federated Learning Neural Tangent Kernel(FL-NTK). Unlike the good property of the symmetric Gram matrix in classical NTK, we show the Gram matrix of FL-NTK is asymmetric in Section~\ref{sec:ntk}. Our techniques address the core question:
\begin{center}
\textit{How shall we handle the asymmetric Gram matrix in FL-NTK?}
\end{center}

\paragraph{Contributions.} Our contributions are summarized into the following two folds:
\begin{itemize}
    \item We proposed a framework to analyze federated learning in neural networks. By appealing to recent advances of over-parameterized neural networks, we prove convergence and generalization results of federated learning without the assumptions on the convexity of objective functions or distribution of data in the convergence analysis. Thus, we make the first step toward bridging the gap between the empirical success of federated learning and its theoretical understanding in the settings of ReLU neural networks. The results theoretically show that given fixed training instances, the number of communication rounds increases as the number of clients increases, which is also supported by empirical evidence.
    We show that when the neural networks are sufficiently wide,
    the training loss across all clients converges to zero at a linear rate.
    Furthermore, we also prove a data-dependent generalization bound.
    \item In federated learning, the update in the global model is no longer determined by the gradient directions directly. Indeed, gradients' heterogeneity in multiple local steps hinders the usage of standard neural tangent kernel analysis, which is based on the kernel gradient descent in the function space for a positive semi-definite kernel. We identify the dynamics of training loss by considering all intermediate states of local steps and establishing the tangent kernel space associated with a general non-symmetric Gram matrix to address this issue. We prove that this Gram matrix is close to symmetric at initialization using concentration properties at initialization. Therefore, we guarantee linear convergence results. This technique may further improve our understanding of many different FL optimization and aggregation methods on neural networks.
\end{itemize}

\paragraph{Organization.} In Section~\ref{sec:relate} we discuss related work. In Section~\ref{sec:prob} we formulate FL convergence problem. In Section~\ref{sec:results} we state our result. In Section~\ref{sec:overview} we summarize our technique overviews. In Section~\ref{sec:proofsketch} we give a proof sketch of our result. In Section \ref{sec:exp}, we conduct experiments that affirmatively support our theoretical results. In Section~\ref{sec:discussion} we conclude this paper and discuss future works.

 %%% Section 1. Introduction
\section{Related Work}
\label{sec:relate}
\paragraph{Federated Learning}
Federated learning has emerged as an important paradigm in distributed deep learning. Generally, federated learning can be achieved by two approaches: 1) each party training the model using private data and where only model parameters being transferred and 2) using encryption techniques to allow safe communications between different parties~\cite{ylct19}. In this way, the details of the data are not disclosed in between each party. In this paper, we focus on the first approach, which has been studied in~\cite{dcm+12, ss15,mmra16,mmr+17}. Federated average (FedAvg)~\cite{mmr+17} firstly addressed the communication efficiency problem by introducing a global model to aggregate local stochastic gradient descent updates. Later, different variations and adaptations have arisen.  This encompasses a myriad of possible approaches, including developing better optimization algorithms~\cite{lsz+20,wys+20} and generalizing model to heterogeneous clients under special assumptions~\cite{zls+18,kma+19,ljz+21}.

Federated learning has been widely used in different fields. Healthcare applications have started to utilize FL for multi-center data learning to solve small data, and privacy in data sharing issues ~\cite{lgd+20,rhl+20,lmx+19,atb+20}. We have also seen new FL algorithms popping up~\cite{wts+19,llh+20,cys+20} in mobile edge computing. FL also has promising applications in autonomous driving~\cite{llc+19}, financial filed~\cite{yzy+19}, and so on.

\paragraph{Convergence of Federated Learning}
Despite its promising benefits, FL comes with new challenges to tackle, especially for its convergence analysis under communication-efficiency algorithms and data heterogeneity. The convergence of the general FL framework on neural networks is underexplored. A recent work~\cite{ljz+21} studies FL convergence on one-layer neural networks. Nevertheless, it is limited by the assumption that each client performs a single local update epoch. Another line of approaches does not directly work on neural network setting~\cite{lhy+19,kmr20,yyz19}. Instead, they make assumptions on the convexity and smoothness of the objective functions, which are not realistic for non-linear neural networks.

 %%% Section 2. Related Work

\section{Problem Formulation as Neural Tangent Kernel}
\label{sec:prob}
To capture the training dynamic of FL on ReLU neural networks, we formulate the problem in Neural Tangent Kernel regime.

\paragraph{Notations}
We use $N$ to denote the number of clients and use $c$ to denote its index. We use $T$ to denote the number of communication rounds, and use $t$ to denote its index. We use $K$ to denote the number of local update steps, and we use $k$ to denote its index. We use $u(t)$ to denote the aggregated server model after round $t$. We use $w_{k,c}(t)$ to denote $c$-th client's model in round $t$ and step $k$.

Let $S_1 \cup S_2 \cup \cdots \cup S_N = [n]$ and $S_i \cap S_j = \emptyset$. 
Given $n$ input data points and labels $\{ (x_1,y_1), (x_2,y_2), \cdots, (x_n,y_n) \}$ in $\R^d \times \R$, the data of each client $c$ is $\{(x_i,y_i):i \in S_c\}$. Let $\phi(z) = \max\{z,0\}$ denote the ReLU activation.

For each client $c \in [N]$, we use $y_c \in \R^{|S_c|}$ to denote the ground truth with regard to its data, and denote $y^{(k)}_c(t) \in \R^{|S_c|}$ to be the (local) model's output of its data in the $t$-th global round and $k$-th local step. For simplicity, we also use $y^{(k)}(t) \in \R^n$ to denote aggregating all (local) model's outputs in the $t$-th global round and $k$-th local step.

\subsection{Preliminaries}

\begin{algorithm*}[h]\caption{Training Neural Network with FedAvg under NTK setting.}\label{alg:alg_main_text}
\begin{algorithmic}[1]
\State $u_r(0) \sim \N(0,I_d)$ for $r\in [m]$. \Comment{$u \in \R^{d \times m}$}
\For{$t = 1, \ldots, T$}
    \For{$c = 1, \ldots, N$}
        \State $w_{0,c}(t) \leftarrow u(t)$ \Comment{$w_{0,c}(t),u(t) \in \R^{d \times m}$}
        \For{$k=1,\ldots,K$}
            \For{$i \in S_c$} 
            \State $y_c^{(k)}(t)_i \leftarrow \frac{1}{ \sqrt{m} } \sum_{r=1}^m a_r \phi(w_{k,c,r}(t)^\top x_i)$ \Comment{$y_c^{(k)}(t) \in \R^{|S_c|}$}
            \EndFor
            \For{$r = 1 \rightarrow m$}
            \For{$i \in S_c$} 
				\State $J_{i,:} \leftarrow \frac{1}{\sqrt{m}}a_r\phi'(w_{k,c,r}(t)^\top x_i)x_i^\top$ \Comment{$J_{i,:} = \frac{\partial f(w_{k,c}(t),x_i,a)}{\partial w_{k,c,r}(t)} \in \R^{1 \times d}$}
			\EndFor
			\State $\text{grad}_r \leftarrow - J^\top (y_{c} - y_c^{(k)}(t) )$
			\Comment{$J \in \R^{|S_c| \times d}$}
            \State $w_{k,c,r}(t) \leftarrow w_{k-1,c,r}(t) - \eta_{\mathrm{local}} \cdot \text{grad}_r $
            \EndFor
        \EndFor
        \State
        $\Delta u_c \leftarrow w_{k,c}(t) - u(t)$
    \EndFor
    \State $\Delta u \leftarrow \frac{1}{N} \sum_{c \in [N]} \Delta u_c$ \Comment{$\Delta u \in \R^{d \times m}$}
    \State $u(t+1) \leftarrow u(t) + \eta_{\mathrm{global}} \Delta u$ \Comment{$u(t+1)\in \R^{d \times m}$}
\EndFor
\end{algorithmic}
\end{algorithm*}

In this subsection we introduce Algorithm \ref{alg:alg_main_text}, the brief algorithm of our federated learning (under NTK setting):

\begin{itemize}
    \item In the $t$-th global round, server broadcasts $u(t) \in \R^{d \times m}$ to every client.
    \item Each client $c$ then starts from $w_{0,c}(t) = u(t)$ and takes $K$ (local) steps gradient descent to find $w_{K,c}(t)$. 
    \item Each client sends $\Delta u_c(t) = w_{K,c}(t)-w_{0,c}(t)$ to server.
    \item Server computes a new $u(t+1)$ based on the average of all $\Delta u_c(t)$. Specifically, the server updates $u(t)$ by the average of all local updates $\Delta u_c(t)$ and arrives at $u(t+1) = u(t) + \eta_\glo \cdot \sum_{c \in [N]} \Delta u_c(t) /N$. 
    \item We repeat the above four steps by $T$ times.
\end{itemize}

\paragraph{Setup}

We define one-hidden layer neural network function $f : \R^d \rightarrow \R$ similar as \cite{dzps19,sy19,bpsw21,sz21}
\begin{align*}
    f (u,x) := \frac{1}{ \sqrt{m} } \sum_{r=1}^m a_r \phi ( u_r^\top x ),
\end{align*}
 where $u \in \R^{d \times m}$ and $u_r \in \R^d$ is the $r$-th column of matrix $u$.
\begin{definition}[Initialization]\label{def:initialization}
We initialize $u \in \R^{d \times m}$ and $a \in \R^m$ as follows:
\begin{itemize}
    \item For each $r \in [m]$, $u_r$ is sampled from ${\cal N}(0,\sigma^2 I)$. 
    \item For each $r \in [m]$, $a_r$ is sampled from $\{-1,+1\}$ uniformly at random (we don't need to train).
\end{itemize}
\end{definition}

We define the loss function for $j \in [N]$,
\begin{align*}
    L_j (u,x): = & ~ \frac{1}{2} \sum_{i \in S_j} ( f(u,x_i) - y_i )^2, \\
    L(u,x) := & ~ \frac{1}{N} \sum_{j=1}^N  L_j(u,x).
\end{align*}

We can compute the gradient $\frac{ \partial f(u,x)}{ \partial u_r } \in \R^d$ (of function $f$),
\begin{align*}
    \frac{ \partial f(u,x)}{ \partial u_r } = \frac{1}{ \sqrt{m} } a_r x {\bf 1}_{ u_r^\top x \geq 0 } .
\end{align*}

We can compute the gradient $ \frac{ \partial L_c (u) }{ \partial u_r } \in \R^d $ (of function $L$),
\begin{align*}
    \frac{ \partial L_c (u) }{ \partial u_r } = \frac{1}{\sqrt{m}} \sum_{i = 1}^n ( f(u,x_i) - y_i) a_r x_i {\bf 1}_{ u_{r,c}^\top x_i \geq 0 } .
\end{align*}

We formalize the problem as minimizing the sum of loss functions over all clients:
\begin{align*}
    \min _{u \in \mathbb{R}^{d \times M}}
    L(u).
\end{align*}

\paragraph{Local update}

In each local step, we update $w_{k,c}$ by gradient descent.

\begin{align*}
    w_{k+1,c} \leftarrow w_{k,c} -  \eta_{\loc}\cdot \frac{ \partial L_c (w_{k,c}) }{ \partial w_{k,c} }.
\end{align*}
Note that 
\begin{align*}
     \frac{ \partial L_c (w_{k,c}) }{ \partial w_{k,c,r} }=  \frac{1}{\sqrt{m}} \sum_{i \in S_c} ( f(w_{k,c},x_i) - y_i) a_r x_i {\bf 1}_{ w_{k,c,r}^\top x_i \geq 0 }.
\end{align*}

\paragraph{Global aggregation}

In each global communication round we aggregate all local updates of clients by taking a simple average
\begin{align*}
    \Delta u(t) = \sum_{c \in [N]}\Delta u_c(t) / N ,
\end{align*}
where $\Delta u_c(t) = w_{K,c} - w_{0,c}$ for all $c \in [N]$.

\paragraph{Global steps in total}

Then global model simply add $\Delta u(t)$ to its parameters.
\begin{align*}
    u(t+1) \leftarrow u(t) + \eta_{\glo} \cdot \Delta u(t).
\end{align*}

\subsection{NTK Analysis}
\label{sec:ntk}

The neural tangent kernel $H^{\infty} \in \R^{n \times n}$, introduced in \cite{jgh18}, is given by 
\begin{align*}
H^{\infty}_{i,j} := & ~ \E_{w \sim \N(0,I)} \left[ x_i^\top x_j {\bf 1}_{ w^\top x_i \geq 0, w^\top x_j \geq 0 } \right]
%= & ~ x_i^\top x_j\big(\pi-\mathrm{arccos}(x_i^\top x_j)\big)/(2\pi).
\end{align*}

At round $t$, let $y(t) = (y_1(t),y_2(t),\cdots,y_n(t)) \in \R^n$ be the prediction vector where $y_i(t) \in \R$ is defined as 
\begin{align*}
    y_i(t) = f(u(t),x_i).
\end{align*}
Recall that we denote labels $y=(y_1,\cdots,y_n) \in \R^n$, $y_c=\{y_i\}_{i \in S_c}$, predictions $y(t) = (y(t)_1,\cdots,y(t)_n)$ and $y_c(t)= \{y(t)_i\}_{i \in S_c}$, we can then rewrite $\|y-y(t)\|_2^2$ as follows:
\begin{align}\label{eq:loss-update}
    & ~ \|y-y(t+1)\|_2^2 \notag\\
    = & ~ \|y-y(t) - (y(t+1)-y(t))\|_2^2 \notag\\
    = & ~ \|y-y(t)\|_2^2 - 2(y-y(t))^\top (y(t+1)-y(t)) 
     ~ +\|y(t+1)-y(t)\|_2^2.
\end{align}

Now we focus on $y(t+1)-y(t)$, notice for each $i \in [n]$
\begin{align}\label{eq:model-update}
&    y_i(t+1)- y_i(t) \notag \\
= & \frac{1}{\sqrt{m}}\sum_{r=1}^{m}a_r(\phi(u_r(t+1)^\top x_i) - \phi ( u_r^\top(t) x ))\notag \\
= & \frac{1}{\sqrt{m}}\sum_{r=1}^{m}a_r\bigg(\phi\big((u_r(t+1)+\eta_{\glo} \Delta u_r(t))^\top x_i\big) -  \phi ( u_r^\top(t) x )\bigg)
\end{align}
where
\begin{align*}
\Delta u_r(t)
:= ~ \frac{a_r}{N}\sum_{c \in [N]}\sum_{k \in [K]} \frac{\eta_{\loc}}{\sqrt{m}}\sum_{j \in S_c}(y_j - y_c^{(k)}(t)_j)x_j{\bf 1} _{w_{k,c,r}(t)^\top x_j \geq 0}.
\end{align*}

In order to further analyze Eq~\eqref{eq:model-update}, we separate neurons into two sets. One set contains neurons with the activation pattern changing over time and another set contains neurons with activation pattern holding the same. Specifically for each $i \in [n]$, we define the set $Q_{i} \subset[m]$ of neurons whose activation pattern is certified to hold the same throughout the algorithm
\begin{align*}
Q_{i}:=\left\{r \in[m]: \forall w \in \mathbb{R}^{d} \text { s.t. } \|w-w_{r}(0)\|_{2} \leq R, ~~
\mathbf{1}_{w_{r}(0)^{\top} x_{i} \geq 0}=\mathbf{1}_{w^{\top} x_{i} \geq 0} \right\},
\end{align*}
and use $\overline{Q}_i$ to denote its complement. Then $y_i(t+1)- y_i(t) = v_{1,i} + v_{2,i}$ where
\begin{equation}\label{eq:v_def}
 \begin{aligned}
    v_{1,i} = &~ \frac{1}{\sqrt{m}}\sum_{r \in Q_i}a_r \bigg(\phi\big((u_r(t)+\eta_{\glo}\Delta u_r(t))^\top x_i\big) - \phi(u_r(t)^\top x_i)\bigg),\\
    v_{2,i} = &~ \frac{1}{\sqrt{m}}\sum_{r \in \ov{Q}_i}a_r \bigg(\phi\big((u_r(t)+\eta_{\glo}\Delta u_r(t))^\top x_i\big)  - \phi(u_r(t)^\top x_i)\bigg).
\end{aligned}   
\end{equation}

The benefit of this procedure is that $v_1$ can be written in closed form
\begin{align*}
    v_{1,i}= 
    \frac{\eta_{\glo}\eta_{\loc}}{Nm}\sum_{k\in [K],c \in [N]}\sum_{j \in S_c}\sum_{r \in Q_i} q_{k,c,j,r},
\end{align*}
where 
\begin{align*}
    q_{k,c,j,r} : = -(y_c^{(k)}(t)_j-y_j) x_i^{\top} x_j{\bf 1}_{w_{k,c,r}(t)^\top x_j,u_r(t)^\top x_i \geq 0} ,
\end{align*}
and $v_2$ is sufficiently small which we will show later.

Now, we extend the NTK analysis to FL. We start with defining the Gram matrix for $f$ as follows. 

\begin{definition}
For any $t \in [0,T],k \in [K],c \in [N]$, we define matrix $H(t,k,c) \in \R^{n \times n}$ as follows 
\begin{align*}
    H(t,k,c)_{i,j} =&~ \frac{1}{m} \sum_{r=1}^{m} x_{i}^\top x_j {\bf 1} _{u_{r}^\top x_i \geq 0,w_{k,c,r}(t)^\top x_j \geq 0},\\
    H(t,k,c)_{i,j}^{\perp} =&~ \frac{1}{m} \sum_{r \in \ov{Q}_i} x_{i}^\top x_j {\bf 1}_{u_{r}^\top x_i \geq 0,w_{k,c,r}(t)^\top x_j \geq 0}.
    %H(s,k,c)_{i,j} = \frac{1}{m} \sum_{r=1}^{m} x_{i}^\top x_j {\bf 1} _{u_{r}^\top x_i \geq 0,w_{r,k,c}^\top x_j \geq 0}
\end{align*}
\end{definition}
This Gram matrix is crucial for the analysis of error dynamics.
When $t=0$ and the width $m$ approaches infinity,
the $H$ matrix becomes the NTK,
and with infinite width,
neural networks just behave like kernel methods with respect to the NTK \cite{adh+19b,lsswy20}.
It turns out that in the finite width case~\cite{ll18,als19_dnn,als19_rnn,dzps19,dllwz19,sy19,os20,hy20,cx20,zpd+20,bpsw21,sz21}, the spectral property of the gram matrix also governs convergence guarantees for neural networks.

We can then decompose $- 2(y-y(t))^\top (y(t+1)-y(t))$ into
\begin{align}\label{eq:loss-inner-product}
     -2(y-y(t))^\top (y(t+1)-y(t)) 
    = &~ -2(y-y(t))^\top  (v_1+v_2)\notag \\
    =&~ -\frac{2\eta_{\glo}\eta_{\loc}}{N}\sum_{i \in [n]}\sum_{k\in [K]}\sum_{c \in [N]}\sum_{j \in S_c} p_{i,k,c,j} \notag \\
    &~ - 2\sum_{i \in [n]}(y_i-y_i(t))v_{2,i}
\end{align}
where 
\begin{align*}
    p_{i,k,c,j}:= ~ (y_i-y_i(t)) \cdot (y_j-y_c^{(k)}(t)_j)  \cdot (H(t,k,c)_{i,j} - H(t,k,c)_{i,j}^{\perp}) .
\end{align*}

Now it remains to analyze Eq~\eqref{eq:loss-update} and Eq~\eqref{eq:loss-inner-product}. Our analysis leverages several key observations in the classical Neural Tangent Kernel theory throughout the learning process:
\begin{itemize}
    \item Weights change lazily, namely
    \begin{align*}
        \|u(t+1)-u(t)\|_2 \leq O(1/n) .
    \end{align*}
    \item Activation patterns remain roughly the same, namely 
    \begin{align*}
    \|H(t,k,c)^{\perp}\|_F \leq O(1),\end{align*}
    and
    \begin{align*}
    \|v_2\|_2 \leq O(\|y-y(t)\|_2).
    \end{align*}
    \item Error controls model updates, namely 
    \begin{align*}
    \|y(t+1)-y(t)\|_2 \leq O(\|y-y(t)\|_2).
    \end{align*}
\end{itemize}

Based on the above observations, we show that the dynamics of federated learning is dominated in the following way
\begin{align*}
    ~ \|y-y(t+1)\|_2^2 \approx  \|y-y(t)\|_2^2 - 2\sum_{k \in [K]}(y-y(t))^\top H(t,k) (y-y^{(k)}(t)),
\end{align*}
where the Gram matrix $H(t,k) \in \R^{n \times n}$ comes from combining the $S_c$ columns of $H(t,k,c)$ for all $c \in [N]$. Since $S_1 \cup S_2 \cup \cdots \cup S_N = [n]$ and $S_i \cap S_j = \emptyset$, every $j \in [n]$ belongs to one unique $S_c$ for some $c$ and $H(t,k)_{i,j} = H(t,k,c)_{i,j}, j \in S_c$.

There are two difficulties lies in the analysis of these dynamics. First, unlike the symmetric Gram matrix in the standard NTK theory for centralized training, our FL framework's Gram matrix is asymmetric. Secondly, model update in each global round is influenced by all intermediate model states of all the clients.

To address these difficulties, we bring in two new techniques to facilitate our understanding of the learning dynamics.
\begin{itemize}
\item First, we generalize Theorem 4.2 in \cite{dzps19} to non-symmetric Gram matrices. We show in Lemma~\ref{lem:gram-matrix-diff} that with good initialization $H(t,k)$ is close to the original Gram matrix $H(0)$, so that model could benefit from a linear learning rate determined by the smallest eigenvalue of $H(0)$.
\item Secondly, we leverage concentration properties at initialization to bound the difference between the errors in local steps and the errors in the global step. Specifically, we can show that $y-y^{(k)}(t) \approx y-y(t)$ for all $k \in [K]$. 
\end{itemize}

 %%% Section 3. Problem Formulation
\section{Our Results}
\label{sec:results}

We first present the main result on the convergence of federated learning in neural networks by the following theorem.
\begin{theorem}[Informal version of Theorem~\ref{thm:fl-ntk-convergence}]\label{thm:main_informal}
Let $m = \Omega( \lambda^{-4} n^4 \log (n/\delta) )$, we iid initialize $u_r(0)$, $a_r$ as Definition~\ref{def:initialization} where $\sigma = 1$. Let $\lambda$ denote $\lambda_{\min}(H(0))$. Let $\kappa$ denote the condition number of $H(0)$. For $N$ clients, for any $\epsilon$, let
\begin{align*}
T=O \Big( \frac{N}{\lambda\eta_{\loc}\eta_{\glo}K}\cdot \log(1/\epsilon) \Big),
\end{align*}
there is an algorithm (FL-NTK) runs in $T$ global steps and each client runs $K$ local steps with choosing 
\begin{align*}
    \eta_{\loc} = O\Big( \frac{\lambda}{\kappa Kn^2} \Big) \mathrm{~~~and~~~} \eta_{\glo} = O(1)
\end{align*} 
and outputs weight $U$ with probability at least $1-\delta$ such that the training loss $L(u,x)$ satisfies
\begin{align*}
    L(u,x)=\frac{1}{2N}\sum_{i=1}^N(f(u,x_i)-y_i)^2 \leq \epsilon .
\end{align*}
\end{theorem}

We note that our theoretical framework is very powerful.
With additional assumptions on the training data distribution and test data distribution, we can also show an upper bound for the generalization error of federated learning in neural networks.
We first introduce a distributional assumption, which is standard in the literature (e.g, see \cite{adh+19a,sy19}).
\begin{definition}[Non-degenerate Data Distribution, Definition 5.1 in \cite{adh+19a}]
A distribution $\mathcal{D}$ over $\mathbb{R}^d\times\mathbb{R}$ is \emph{$(\lambda,\delta,n)$-non-degenerate},
if with probability at least $1-\delta$,
for $n$ i.i.d. samples $(x_i,y_i)_{i=1}^n$ chosen from $\mathcal{D}$,
$\lambda_{\min}(H(0))\geq \lambda>0$.
\end{definition}

Our result on generalization bound is stated in the following theorem.
\begin{theorem}[Informal, see Appendix \ref{sec:generalization} for details]
\label{thm:generalization_main_informal}
Fix failure probability $\delta \in (0,1)$.
Suppose the training data $S=\{(x_i,y_i)\}_{i=1}^n$ are i.i.d samples from a $(\lambda,\delta/3,n)$-non-degenerate distribution $\mathcal{D}$,
and 
\begin{itemize} 
\item $\sigma\leq O({\lambda \cdot \poly(\log n,\log(1/\delta))}/{n})$,
\item 
$m\geq \Omega ( \sigma^{-2} \cdot \poly(n,\log m,\log(1/\delta),\lambda^{-1}) )$,
\item $T\geq \Omega ( ( \eta_{\loc}\eta_{\glo}K\lambda )^{-1} N \log(n/\delta) )$ .
\end{itemize}
We initialize $u \in \R^{d \times m }$ and $a \in \R^m$ as Definition~\ref{def:initialization}. 
Consider any loss function $\ell:\mathbb{R}\times \mathbb{R}\rightarrow [0,1]$ that is 1-Lipschitz in its first argument.
Then with probability at least $1-\delta$ the following event happens: after $T$  global steps,  the generalization loss 
\begin{align*}
    L_{\mathcal{D}}(f) := \E_{(x,y)\sim \mathcal{D}}[\ell(f(u,x),y)]
\end{align*}
is upper bounded as
\begin{align*}
L_{\mathcal{D}}(f)\leq & ~ ( {2y^\top (H^{\infty})^{-1}y}/{n} )^{1/2} +O (\sqrt{ \log(n / (\lambda \delta) ) / n} ).
\end{align*}
\end{theorem}

\section{Techniques Overview}
\label{sec:overview}
In our algorithm,
when $K=1$, i.e., we only perform one local step per global step,
essentially we are performing gradient descent on all the $n$ data points with step size ${\eta_{\mathrm{global}}\eta_{\mathrm{local}} }/{N}$.
As the norm of the gradient is proportional to $1/\sqrt{m}$,
when the neural network is sufficiently wide,
we can control the norm of the gradient.
Then by the update rule of gradient descent, we hence upper bound the movement of the first layer weight $u$.
By anti-concentration of the normal distribution, this implies that for each input $x$, the activation status of most of the  ReLU gates remains the same as initialized, which enables us to apply the standard convergence analysis of gradient descent on convex and smooth functions.
Finally, we can handle the effect of ReLU gates whose activation status have changed by carefully choosing the step size.

However, the analysis becomes much more complicated when $K\geq 2$,
where the movement of $u$ is no longer determined by the gradient directly.
Nevertheless,
on each client, we are still performing gradient descent for $K$ local steps.
So we can handle the movement of the local weight $w$.
The major technical bulk of this work is then proving that the training error shrinks when we set the global weight movement as the average of the local weight.
Our argument is inspired by that of \cite{dzps19}
but is much more involved.

\section{Proof Sketch}
\label{sec:proofsketch}

In this section we sketch our proof of  Theorem \ref{thm:main_informal} and Theorem~\ref{thm:generalization_main_informal}. The detailed proof is deferred to Appendix \ref{sec:full_proof}.

In order to prove the linear convergence rate in 
Theorem \ref{thm:main_informal},
it is sufficient to show that the training loss shrinks in each round, or formally for each $\tau=0,1,\cdots$,
\begin{equation}\label{eq:main_induction}
   \begin{aligned}
    \| y (\tau+1) - y \|_2^2 
    \leq & \Big( 1 - \frac{\lambda\eta_{\glo} \eta_{\loc}  K}{2N} \Big) \cdot \| y (\tau) - y \|_2^2.
\end{aligned} 
\end{equation}
We prove Eq.~\eqref{eq:main_induction} by induction.
Assume that we have proved for $\tau\leq t-1$ and we want to prove Eq.~\eqref{eq:main_induction} for $\tau = t$.
We first show that the movement of the weight $u$ is bounded under the induction hypothesis.
\begin{lemma}[Movement of global weight, informal version of Lemma \ref{lem:bound-global-weights}]\label{lem:global_weights_informal}
For any $r\in [m]$,
\begin{align*}
    \| u_r(t) - u_r(0) \|_2 = O \Big( \frac{ \sqrt{n} \| y - y (0) \|_2 }{ \sqrt{m} \lambda } \Big).
\end{align*}
\end{lemma}
The detailed proof can be found in Appendix \ref{sec:appendix_tech}.

We then turn to the proof of Eq.~\eqref{eq:main_induction} by decomposing the loss in $t+1$-th global round
\begin{align}\label{eq:decomposition-loss}
    & ~ \|y-y(t+1)\|_2^2 \notag \\
    = & ~ \|y-y(t)\|_2^2 - 2(y-y(t))^\top (y(t+1)-y(t))% \notag \\
    %& ~ 
    +\|y(t+1)-y(t)\|_2^2 %\notag \\
   % = & ~ \|y-y(t)\|_2^2 + C_1 + C_2 + C_3 + C_4.
\end{align}
To this end, we need to investigate the change of prediction in consecutive rounds, which is described in Eq.~\eqref{eq:model-update}.
For the sake of simplicity,
we introduce the notation
\begin{align*}
    z_{i,r} :=  \phi\big((u_r(t)+\eta_{\glo}\Delta u_r(t))^\top x_i\big) - \phi(u_r(t)^\top x_i),
\end{align*}
then we have
\begin{align*}
     y(t+1)_i-y(t)_i
    = & ~ \frac{1}{\sqrt{m}}\sum_{r =1}^m a_r z_{i,r} 
     =  ~\frac{1}{\sqrt{m}}\sum_{r \in Q_i}a_r z_{i,r} + v_{2,i},
\end{align*}
where $v_{2,i}$ is introduced in Eq.~\eqref{eq:v_def}.

For client $c \in [N]$ let $y_c^{(k)}(t)_j$ ($j \in S_c$) be defined by
$  y_c^{(k)}(t)_j = f(w_{k,c}(t),x_j)$.
By the gradient-averaging scheme described in Algorithm~\ref{alg:alg_main_text},
$\Delta u_r(t)$, the change in the global weights is
\begin{align*}
    \frac{a_r}{N}\sum_{c \in [N]}\sum_{k \in [K]} \frac{\eta_{\loc}}{\sqrt{m}}\sum_{j \in S_c}(y_j - y_c^{(k)}(t)_j)x_j{\bf 1} _{w_{k,c,r}(t)^\top x_j \geq 0}.
\end{align*}
Therefore, we can calculate $\frac{1}{\sqrt{m}}\sum_{r \in Q_i}a_r z_{i,r}$ 
\begin{align*}
&~\frac{1}{\sqrt{m}}\sum_{r \in Q_i}a_r z_{i,r} \\
   = &~ \frac{1}{\sqrt{m}}\sum_{r \in Q_i}a_r \bigg(\phi\big((u_r(t)+\eta_{\glo}\Delta u_r(t))^\top x_i\big) - ~\phi(u_r(t)^\top x_i)\bigg)\\
    =&~ \frac{\eta_{\glo}\eta_{\loc}}{m N}\sum_{k\in [K]}\sum_{c \in [N]}\sum_{j \in S_c}(y_j-y_c^{(k)}(t)_j)x_i^\top x_j
    ~ \cdot \sum_{r \in Q_i}{\bf 1}_{u_{r}^\top x_i \geq 0,w_{k,c,r}(t)^\top x_j \geq 0}\\
    =&~ \frac{\eta_{\glo}\eta_{\loc}}{N}\sum_{k\in [K]}\sum_{c \in [N]}\sum_{j \in S_c} 
     ~ (y_j-y_c^{(k)}(t)_j) (H(t,k,c)_{i,j} - H(t,k,c)_{i,j}^{\perp}).
\end{align*}
Further, we write $-2(y-y(t))^\top (y(t+1)-y(t))$ as follows:
\begin{align*}
    & ~ -2(y-y(t))^\top (y(t+1)-y(t))  \\
    = &~ -2(y-y(t))^\top  (v_1+v_2)\\
    =&~ -\frac{2\eta_{\glo}\eta_{\loc}}{N}\sum_{i \in [n]}\sum_{k\in [K]}\sum_{c \in [N]}\sum_{j \in S_c}  (y_i-y_i(t))(y_j-y_c^{(k)}(t)_j)
      (H(t,k,c)_{i,j} - H(t,k,c)_{i,j}^{\perp})\\
    &~ - 2\sum_{i \in [n]}(y_i-y_i(t))v_{2,i}.
\end{align*}

To summarize, we can decompose the loss as
\begin{align*}
    ~ \|y-y(t+1)\|_2^2 
   =  ~ \|y-y(t)\|_2^2 + C_1 + C_2 + C_3 + C_4.
\end{align*}
where
\begin{align*}
    C_1 =&~ -\frac{2\eta_{\glo}\eta_{\loc}}{N}\sum_{i \in [n]}\sum_{k\in [K]}\sum_{c \in [N]}\sum_{j \in S_c}
    ~ (y_i-y_i(t))(y_j-y_c^{(k)}(t)_j) H(t,k,c)_{i,j},\\
    C_2 =&~ + \frac{2\eta_{\glo}\eta_{\loc}}{N}\sum_{i \in [n]}\sum_{k\in [K]}\sum_{c \in [N]}\sum_{j \in S_c}
    ~ (y_i-y_i(t))(y_j-y_c^{(k)}(t)_j) H(t,k,c)_{i,j}^{\perp},\\
    C_3 =&~ - 2\sum_{i \in [n]}(y_i-y_i(t))v_{2,i},\\
    C_4 =&~ + \|y(t+1)-y(t)\|_2^2.
\end{align*}

Let $R = \max_{r \in [m]}  \| u_r(t) - u_r(0) \|_2$ be the maximal movement of global weights. Note that by Lemma \ref{lem:global_weights_informal},
$R$ can be made arbitrarily small as long as the width $m$ is sufficiently large.
Next, we bound $C_1,C_2,C_3,C_4$, by arguing that they are bounded from above if $R$ is small and the learning rate is properly chosen.
The detailed proof is deferred to Appendix \ref{sec:C_upper_bounds}.
\begin{lemma}[Bounding of $C_1,C_2,C_3,C_4$, informal version of Claim~\ref{cla:bound_C1}, Claim~\ref{cla:bound_C2}, Claim~\ref{cla:bound_C3} and Claim~\ref{cla:bound_C4}]
Assume that
\begin{itemize}
\item $R = O(\lambda/n)$, 
\item $\eta_{\loc} = O(1/(\kappa Kn^2))$, 
\item $\eta_{\glo} = O(1)$.
\end{itemize}
Then with high probability we have
\begin{align*}
    C_1\leq & ~ -\eta_{\glo}\eta_{\loc}\lambda K\|y-y(t)\|_2^2/N,\\
    C_2 \leq & ~ + \eta_{\glo}\eta_{\loc}\lambda K\|y-y(t)\|_2^2/(40N),\\
    C_3 \leq & ~ + \eta_{\glo}\eta_{\loc}\lambda K\|y-y(t)\|_2^2/(40N),\\
    C_4 \leq & ~ + \eta_{\glo}\eta_{\loc}\lambda K\|y-y(t)\|_2^2/(40N).
\end{align*}
\end{lemma}

Combining the above lemma,
we arrive to 
\begin{align*}
\|y-y(t+1)\|_2^2 \leq \Big( 1 - \frac{\eta_{\glo}\eta_{\loc}\lambda K}{2N} \Big) \cdot\|y-y(t)\|_2^2 ,
\end{align*}
which completes the proof
of Eq.~\eqref{eq:main_induction}.
Finally Theorem \ref{thm:main_informal} follows from 
\begin{align*}
\|y - y(T)\|_2^2 \leq \Big( 1 - \frac{\eta_{\glo}\eta_{\loc}\lambda K}{2N} \Big)^T \leq \epsilon.
\end{align*}

 %%%Section 5. technique overview
\section{Experiment}
\label{sec:exp}

\begin{figure*}[h]
    \captionsetup[subfigure]{justification=centering}
    \centering
    \subfloat[Local update epoch $K$ =2 ]{\includegraphics[width=0.48\linewidth]{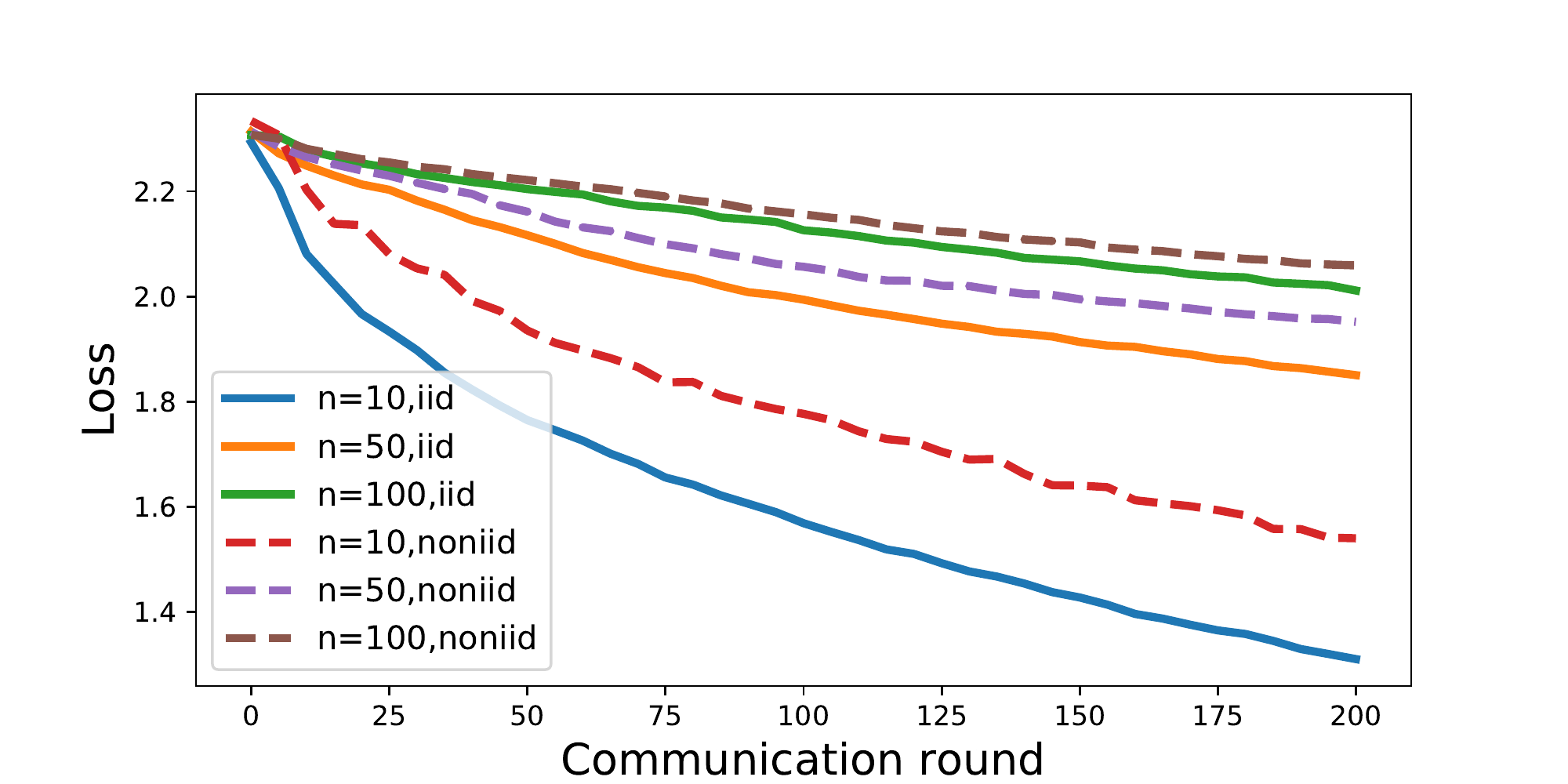}}
    ~
    \subfloat[Local update epoch $K$ =5]{\includegraphics[width=0.48\linewidth]{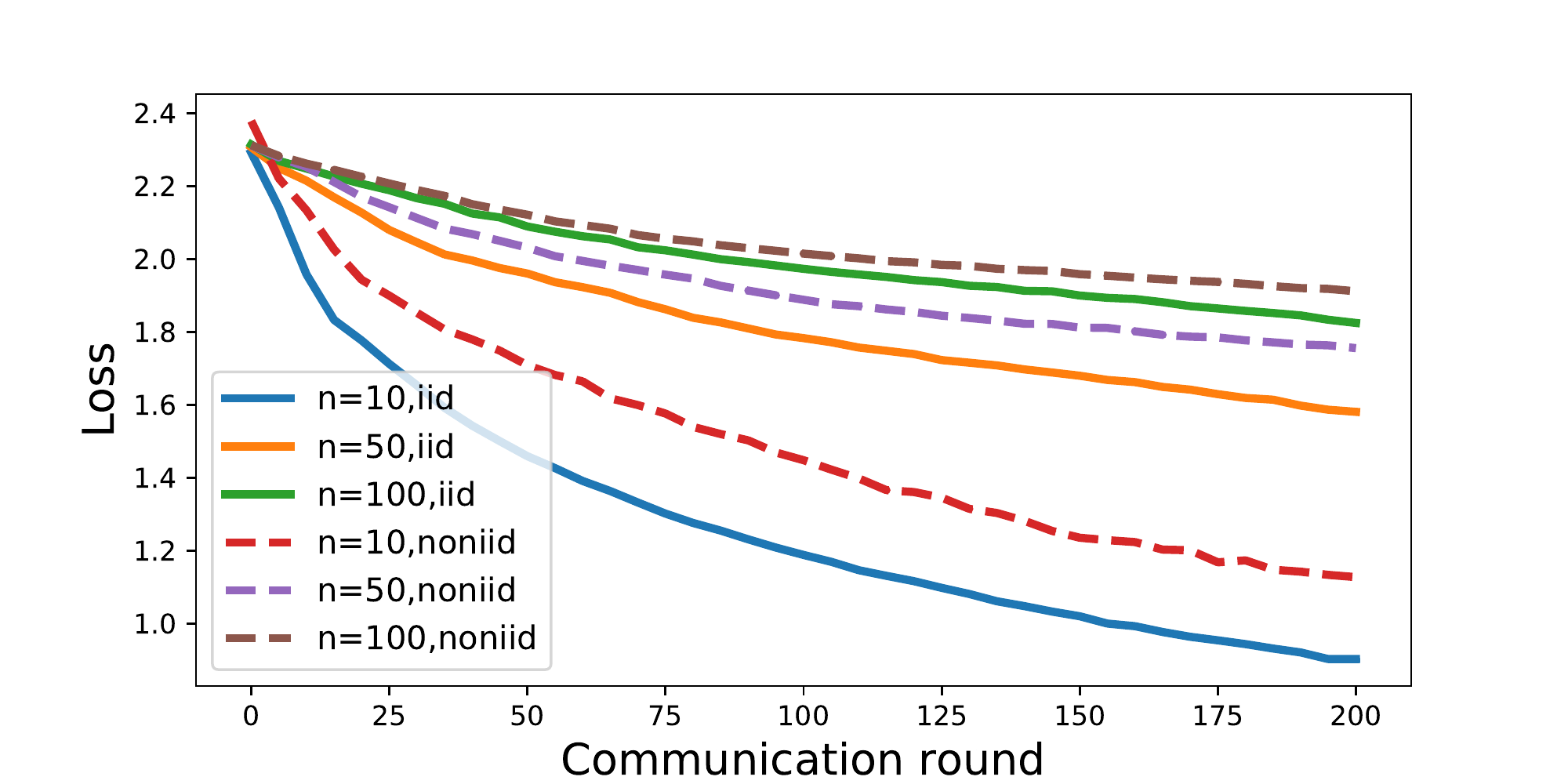}}
    \caption{Training loss vs. communication rounds when number of clients $N=10, 50, 100$ with iid and non-iid setting using mini-batch SGD optimizer.}
    \label{fig:cifar}
\end{figure*}

\paragraph{Models and datasets} We examine our theoretical results on a benchmark dataset - Cifar10 in FL study. We perform 10 class classification tasks using ResNet56 \cite{hzr+16}. For fair convergence comparison, \textit{we fixed the total number of samples $n$}. Based on our main result Theorem~\ref{thm:main_informal}, we show the convergence with respect to the number of client $N$. To clearly evaluate the effects on $N$, we set all the clients to be activated (sampling all the clients) at each aggregation. We examine the settings 
of both non-iid and iid clients: 
\begin{itemize}
    \item[\textbf{iid}] Data distribution is homogeneous in all the clients. Specifically, the label distribution over 10 classes is a uniform distribution. 
\item[\textbf{non-iid}] Data distribution is heterogeneous in all the clients. For non-IID splits, we utilize the Dirichlet distribution as in~\cite{yagg+19,wys+20,haa20}. First, each of these datasets is heterogeneously divided into $J$
batches. The heterogeneous partition allocates $p_z \sim \rm{Dir}_J(
\alpha)$ proportion of the instances of label $z$ to batch $j$. Then one label is sampled based on these vectors for each device, and an image is sampled without replacement based on the label. For all the classes, we repeat this process until all data points are assigned to devices. 
\end{itemize}

\paragraph{Setup} The experiment is conducted on one Ti2080 Nvidia GPU. We use SGD with a learning rate of 0.03 to train the neural network for 160 communication rounds\footnote{It is difficult to run real NTK experiment or full-batch gradient descent (GD) due to memory contrains.}. We set batch size as 128. We set the local update epoch $K=2$ and $K=5$\footnote{Local update \textit{step} equals to \textit{epoch} in GD. But the number of \textit{steps} of \textbf{one} \textit{epoch} in SGD is equal to the number of mini-batches, $J$.}, and Dirichlet distribution parameter $\alpha=0.5$. There are 50,000 training instances in Cifar10. We vary $N = 10, 50, 100$ and record the training loss over communication rounds. The implementation is based on FedML~\cite{hls+20}.

\paragraph{Impact of $N$} Our theory suggests that given fixed training instances (fixed $n$) a smaller $N$ requires less communication rounds to converge to a given $\epsilon$. In other words, a smaller $N$ may accelerate the convergence of the global model. Intuitively, a large $N$ on a fixed amount of data means a heavier communication burden. Figure~\ref{fig:cifar} shows the training loss curve of different choices of $N$. We empirically observe that a smaller $N$ converges faster given a fix number of total training samples, which is affirmative to our theoretical results.  

 %%%% Section 7. Experiment
\section{Discussion}
\label{sec:discussion}
Overall, we provide the first comprehensive proof of convergence of gradient descent and generalization bound for over-parameterized ReLU neural network in federated learning. We consider the training dynamic with respect to local update steps $K$ and number of clients $N$. 

Different from most of existing theoretical analysis work on FL, FL-NTK prevails over the ability to exploit neural network parameters. There is a great potential for FL-NTK to understand the behavior of different neural network architectures, i.e., how Batch Normalization affects convergence in FL, like FedBN~\cite{ljz+21}. Different from FedBN, whose analysis is limited to $K=1$, we provide the first general framework for FL convergence analysis by considering $K \geq 2$. The extension is non-trivial, as parameter update does not follow the gradient direction due to the heterogeneity of local data. To tackle this issue, we establish the training trajectory by considering all intermediate states and establishing an asymmetric Gram matrix related to local gradient aggregations. We show that with quartic network width, federated learning can converge to a global-optimal at a linear rate. We also provide a data-dependent generalization bound of over-parameterized neural networks trained with federated learning. 

It will be interesting to extend the current FL-NTK framework for multi-layer cases. Existing NTK results of two-layer neural networks~(NNs) have shown to be generalized to multi-layer NNs with various structures such as RNN, CNN, ResNet, etc. \cite{als19_dnn,als19_rnn,dllwz19}. The key techniques for analyzing FL on wide neural networks are addressed in our work. Our result can be generalized to multi-layer NNs by controlling the perturbation propagation through layers using well-developed tools (rank-one perturbation analysis, randomness decomposition, and extended McDiarmid’s Inequality). We hope our results and techniques will provide insights for further study of distributed learning and other optimization algorithms.
 %%%Section 8. Discussion

\section*{Acknowledgements}

Baihe Huang is supported by the Elite Undergraduate Training Program of School of Mathematical Sciences at Peking University, and work done while interning at Princeton University and Institute for Advanced Study (advised by Zhao Song). Xin Yang is supported by NSF grant CCF-2006359. This project is partially done while Xin was a Ph.D. student at University of Washington. The authors would like to thank the ICML reviewers for their valuable comments.

%\begin{comment}
\newpage
\appendix 
\onecolumn
\newpage
\section*{Appendix}

\paragraph{Roadmap:}
In Appendix~\ref{sec:tools}, we list several probability results. In Appendix~\ref{sec:full_proof} we prove our convergence result of FL-NTK. In Appendix~\ref{sec:generalization}, we prove our generalization result of FL-NTK.

\section{Probability Tools}\label{sec:tools}

\begin{lemma}[Bernstein inequality \cite{b24}]\label{lem:bernstein}
Let $X_1,\dots,X_n$ be independent zero-mean random variables. Suppose that $|X_i| \leq M$ almost surely, for all $i\in [n]$. Then, for all positive $t $,
\begin{align*}
    \Pr \left[ \sum_{i = 1}^n X_i \leq t \right] \leq \exp \left( -\frac{ t^2 /2 }{ \sum_{j = 1}^n \E[X_j^2] + Mt /3 } \right).
\end{align*}
\end{lemma}

\begin{lemma}[Anti-concentration inequality of Gaussian] \label{lem:anti_gaussian}
Let $X \sim N(0,\sigma^2)$, then for any $0<t \leq \sigma$
\begin{align*}
    \Pr \left[ |X| \leq t \right] \in \big( \frac{2t}{3\sigma}, \sqrt{\frac{2}{\pi}} \cdot \frac{t}{\sigma}\big).
\end{align*}
\end{lemma}

\begin{proof}
For completeness, we provide a short proof. %It follows from the CDF of $\chi^2$ distribution and some properties of the gamma function. 
Since $X\sim N(0,\delta^2)$, the CDF of $X^2$ is $\Pr[X^2 \leq t^2 ] = \frac{\gamma(1/2, t^2/2\sigma^2)}{\Gamma(1/2)}$ where $\gamma(\cdot, \cdot)$ is the incomplete lower gamma function. This can be further simplified to
$\Pr[X^2 \leq t^2 ] = \mathrm{erf}(\sqrt{t^2/2\sigma^2})$ where erf is the error function. For $z\leq 1$, we can sandwich the erf function by $2z/3 \leq \mathrm{erf}(z/\sqrt{2}) \leq \sqrt{2/\pi} z$, thus letting $z = t/\sigma$ complete the proof.

\end{proof}
%!TEX root=main.tex

\section{Convergence of Neural Networks in Federated Learning}\label{sec:full_proof}

\begin{definition}
We let $\kappa$ to denote the condition number of Gram matrix $H(0)$.

\begin{assumption}\label{ass:data_dependent_assumption}
We assume $\|x_i\|_2 = 1$ and $\lambda = \lambda_{\min}(H(0)) \in (0,1]$.
\end{assumption}

\end{definition}
\begin{table}[h]
    \centering
    \begin{tabular}{|l|l|l|} \hline
        {\bf Notation} & {\bf Dimension} & {\bf Meaning} \\ \hline
        $N$ & $\mathbb{N}$ & \#clients \\ \hline
        $c$ & $[N]$ & its index \\ \hline
        $T$ & $\mathbb{N}$ & \#communication rounds \\ \hline
        $t$ & $[T]$ & its index \\ \hline
        $K$ & $\mathbb{N}$ & \#local update steps  \\ \hline
        $k$ & $[K]$ & its index \\ \hline
        $y(t)$ & $ \R^{n}$ & aggregated server model after global round $t$ \\ \hline
        $y_c $ & $ \R^{|S_c|}$ & ground truth of $c$-th client\\ \hline
        $y_c^{(k)}(t) $ & $ \R^{|S_c|}$ & $c$-th client's model in global round $t$ and local step $k$  \\ \hline
        $y^{(k)}(t) $ & $ \R^{n}$ & all client's model in global round $t$ and local step $k$  \\ \hline
        $w_{k,c}(t)$ & $ \R^{d \times m}$ & $c$-th client's model parameter in global round $t$ and local step $k$ \\ \hline
        $u(t) $ & $ \R^{d \times m}$ & aggregated server model parameter in global round $t$ and local step $k$ \\ \hline
    \end{tabular}
    \caption{Summary of several notations}
    \label{tab:summary_of_our_fl_notations}
\end{table}

\subsection{Convergence Result}

\begin{theorem}\label{thm:fl-ntk-convergence}
Recall that $\lambda=\lambda_{\min}(H(0))>0$.
Let $m = \Omega( \lambda^{-4} n^4 \log (n/\delta) )$, we iid initialize $u_r(0) \sim {\N}(0,I)$, $a_r$ sampled from $\{-1,+1\}$ uniformly at random for $r\in [m]$, and we set the step size $\eta_\loc = O( \lambda / (\kappa Kn^2) )$ and $\eta_\glo = O(1)$, then with probability at least $1-\delta$ over the random initialization we have for $t = 0,1,2,\cdots$
\begin{align}\label{eq:fl-global}
\| y (t) - y \|_2^2 \leq ( 1 - \frac{\eta_{\glo} \eta_{\loc} \lambda K}{2N} )^t \cdot \| y (0) - y \|_2^2.
\end{align}
\end{theorem}

\begin{proof}
We prove by induction. The base case is $t=0$ and it is trivially true.
Assume for $\tau=0,\cdots,t$ we have proved Eq.~\eqref{eq:fl-global} to be true. We show Eq.~\eqref{eq:fl-global} holds for $\tau=t+1$.

Recall that the set $Q_{i} \subset[m]$ is defined as follow
\begin{align*}
Q_{i}:=\left\{r \in[m]: \forall w \in \mathbb{R}^{d} \text { s.t. } \|w-w_{r}(0)\|_{2} \leq R, ~~
\mathbf{1}_{w_{r}(0)^{\top} x_{i} \geq 0}=\mathbf{1}_{w^{\top} x_{i} \geq 0} \right\},
\end{align*}
and $\overline{Q}_i$ denotes its complement.

Let $v_{1,i}, v_{2,i}$ be defined as follows
\begin{align*}
    v_{1,i} = &~ \frac{1}{\sqrt{m}}\sum_{r \in Q_i}a_r \bigg(\phi\big((u_r(t)+\eta_{\glo}\Delta u_r(t))^\top x_i\big) - \phi(u_r(t)^\top x_i)\bigg),\\
    v_{2,i} = &~ \frac{1}{\sqrt{m}}\sum_{r \in \ov{Q}_i}a_r \bigg(\phi\big((u_r(t)+\eta_{\glo}\Delta u_r(t))^\top x_i\big) - \phi(u_r(t)^\top x_i)\bigg).
\end{align*}

Let $H(t,k,c)_{i,j},H(t,k,c)_{i,j}^{\perp}$ be defined as follows
\begin{align*}
     H(t,k,c)_{i,j} =&~ \frac{1}{m} \sum_{r=1}^{m} x_{i}^\top x_j {\bf 1} _{u_{r}^\top x_i \geq 0,w_{k,c,r}(t)^\top x_j \geq 0},\\
     H(t,k,c)_{i,j}^{\perp} =&~ \frac{1}{m} \sum_{r \in \ov{Q}_i} x_{i}^\top x_j {\bf 1}_{u_{r}^\top x_i \geq 0,w_{k,c,r}(t)^\top x_j \geq 0}.
\end{align*}
Define $H(t)$ and $H(t)^{\perp} \in \R^{n \times n}$ as 
\begin{align*}
H(t)_{i, j}=\frac{1}{m} \sum_{r=1}^{m} x_{i}^{\top} x_{j} \mathbf{1}_{u_{r}(t)^{\top} x_{i} \geq 0, u_{r}(t)^{\top} x_{j} \geq 0}, \\
H(t)_{i, j}^{\perp}=\frac{1}{m} \sum_{r \in \bar{Q}_{i}} x_{i}^{\top} x_{j} \mathbf{1}_{u_{r}(t)^{\top} x_{i} \geq 0, u_{r}(t)^{\top} x_{j} \geq 0}.
\end{align*}

Let $y_c^{(k)}(t)_j$ ($j \in S_c$) be defined by
\begin{align*}
    y_c^{(k)}(t)_j = f(w_{k,c}(t),x_j).
\end{align*}

We can write $\Delta u_r(t)$ as follow
\begin{align*}
    \Delta u_r(t) = \frac{a_r}{N}\sum_{c \in [N]}\sum_{k \in [K]} \frac{\eta_{\loc}}{\sqrt{m}}\sum_{j \in S_c}(y_j - y_c^{(k)}(t)_j)x_j{\bf 1} _{w_{k,c,r}(t)^\top x_j \geq 0}.
\end{align*}

Thus we have
\begin{align*}
    v_{1,i} =&~ \frac{\eta_{\glo}\eta_{\loc}}{mN}\sum_{k\in [K]}\sum_{c \in [N]}\sum_{j \in S_c}(y_j-y_c^{(k)}(t)_j)x_i^\top x_j \sum_{r \in Q_i}{\bf 1}_{u_{r}^\top x_i \geq 0,w_{k,c,r}(t)^\top x_j \geq 0}\\
    =&~ \frac{\eta_{\glo}\eta_{\loc}}{N}\sum_{k\in [K]}\sum_{c \in [N]}\sum_{j \in S_c}(y_j-y_c^{(k)}(t)_j) (H(t,k,c)_{i,j} - H(t,k,c)_{i,j}^{\perp}).
\end{align*}

We can therefore write $-2(y-y(t))^\top (y(t+1)-y(t))$ as follow
\begin{align*}
    & ~ -2(y-y(t))^\top (y(t+1)-y(t))  \\
    = &~ -2(y-y(t))^\top  (v_1+v_2)\\
    =&~ -\frac{2\eta_{\glo}\eta_{\loc}}{N}\sum_{i \in [n]}\sum_{k\in [K]}\sum_{c \in [N]}\sum_{j \in S_c}(y_i-y_i(t))(y_j-y_c^{(k)}(t)_j) (H(t,k,c)_{i,j} - H(t,k,c)_{i,j}^{\perp})\\
    &~ - 2\sum_{i \in [n]}(y_i-y_i(t))v_{2,i}.
\end{align*}

Let
\begin{align*}
    C_1 =&~ -\frac{2\eta_{\glo}\eta_{\loc}}{N}\sum_{i \in [n]}\sum_{k\in [K]}\sum_{c \in [N]}\sum_{j \in S_c}(y_i-y_i(t))(y_j-y_c^{(k)}(t)_j) H(t,k,c)_{i,j}\\
    C_2 =&~ \frac{2\eta_{\glo}\eta_{\loc}}{N}\sum_{i \in [n]}\sum_{k\in [K]}\sum_{c \in [N]}\sum_{j \in S_c}(y_i-y_i(t))(y_j-y_c^{(k)}(t)_j) H(t,k,c)_{i,j}^{\perp}\\
    C_3 =&~ - 2\sum_{i \in [n]}(y_i-y_i(t))v_{2,i}\\
    C_4 =&~ \|y(t+1)-y(t)\|_2^2.
\end{align*}

Then 
\begin{align*}
    & ~ \|y-y(t+1)\|_2^2 \\
    = & ~ \|y-y(t)\|_2^2 - 2(y-y(t))^\top (y(t+1)-y(t)) +\|y(t+1)-y(t)\|_2^2\\
    =&~ \|y-y(t)\|_2^2 + C_1 + C_2 + C_3 + C_4.
\end{align*}

By Claim~\ref{cla:bound_C1}, Claim~\ref{cla:bound_C2}, Claim~\ref{cla:bound_C3} and Claim~\ref{cla:bound_C4} we have
\begin{align*}
    \|y-y(t+1)\|_2^2 \leq&~ \frac{2\eta_{\glo}\eta_{\loc}}{N}(-K\lambda+4nRK(1+2\eta_{\loc}Kn)+2\eta_{\loc}\kappa\lambda K^2n)\|y-y(y)\|_2^2\\
    &~ + \frac{16\eta_{\glo}\eta_{\loc}}{N}K(1+2\eta_{\loc}Kn)nR\|y-y(y)\|_2^2 \\
    &~ +  \frac{16\eta_{\glo}\eta_{\loc}}{N}K(1+2\eta_{\loc}Kn)nR\|y-y(y)\|_2^2\\
    &~ + \frac{4\eta_{\glo}^2\eta_{\loc}^2n^2K^2(1+2\eta_{\loc}Kn)^2}{N^2}\|y-y(y)\|_2^2.
\end{align*}
By the choice of $\eta_{\loc} \leq \frac{\lambda}{1000\kappa n^2 K}$ and $\eta_{\loc} \eta_{\glo} \leq \frac{\lambda}{1000\kappa n^2 K}$ and $R \leq \lambda / (1000 n)$ we come to
\begin{align}\label{eq:convergence}
    \|y-y(t+1)\|_2^2
    \leq & ~ \| y - y(t) \|_2^2 \notag \\
    & ~ - \frac{\eta_{\glo} \eta_{\loc} \lambda K}{N} \|y-y(t)\|_2^2  \\
    & ~ + 40 \frac{ \eta_{\glo}\eta_{\loc}K n R}{N}\|y-y(t)\|_2^2 \notag\\
    & ~ + 40 \frac{ \eta_{\glo}\eta_{\loc}K n R}{N}\|y-y(t)\|_2^2 \notag \\
    & ~ + \frac{\eta_{\loc}^2\eta_{\glo}^2n^2K^2}{N^2}\|y-y(t)\|^2_2 \notag\\
    \leq & ~ \| y - y(t) \|_2^2 \notag\\
    & ~ - (1-1/10)\frac{\eta_{\glo} \eta_{\loc} \lambda K}{N} \|y-y(t)\|_2^2  \notag\\
    & ~ + 80 \frac{ \eta_{\glo}\eta_{\loc}K n R }{N}\|y-y(t)\|_2^2 \notag\\
    \leq & \| y - y(t) \|_2^2 - \frac{1}{2} \frac{\eta_{\glo} \eta_{\loc} \lambda K}{N} \|y-y(t)\|_2^2
\end{align}
where the second step follows from $\eta_{\loc} \leq \frac{\lambda}{1000\kappa n^2 K}$, the third step follows from $R \leq \lambda / (1000 n)$.
\end{proof}

\subsection{Bounding $C_1,C_2,C_3,C_4$}\label{sec:C_upper_bounds}
\begin{claim}\label{cla:bound_C1}
We have with probability at least $1-n^2\cdot \exp(-mR/10)$ over random initialization
\begin{align*}
C_1 \leq \frac{2\eta_{\glo}\eta_{\loc}}{N}\|y-y(t)\|_2^2(-K\lambda + 4nRK(1+2 \eta_{\loc} K n) +  2\eta_{\loc}  \kappa\lambda  K^2 n ).
\end{align*}
\iffalse
Further, if 
\begin{align*}
\eta_{\loc} \leq  1 / (10 K \sqrt{n \kappa}) \mathrm{~~~and,~~~} R \leq \lambda / (10n)
\end{align*}
we have
\begin{align*}
C_1 \leq - \frac{ \eta_{ \glo } \eta_{ \loc  } \lambda K }{ N } \| y - y ( t ) \|_2^2
\end{align*}
\fi
\end{claim}
\begin{proof}
We first calculate 
\begin{align*}
    &~ \sum_{i \in [n]}\sum_{k\in [K]}\sum_{c \in [N]}\sum_{j \in S_c}(y_i-y_i(t))(y_j-y_c^{(k)}(t)_j) H(t,k,c)_{i,j}\\
    =&~ \sum_{i \in [n]}\sum_{k\in [K]}\sum_{c \in [N]}\sum_{j \in S_c}(y_i-y_i(t))(y_j-y_c^{(k)}(t)_j) (H(t,k,c)_{i,j}-H(0)_{i,j})\\
    &~ + \sum_{i \in [n]}\sum_{k\in [K]}\sum_{c \in [N]}\sum_{j \in S_c}(y_i-y_i(t))(y_j(t)-y_c^{(k)}(t)_j) H(0)_{i,j}\\
    &~ + K\sum_{i \in [n]}\sum_{j \in [n]}(y_i-y_i(t))(y_j-y_j(t)) H(0)_{i,j}.
\end{align*}

From Lemma~\ref{lem:bound-global-weights} and Lemma~\ref{lem:fl-local} we have $\|u_r(t)-u(0)\|_2 \leq R$ and $\|w_{k,c,r}(t)-u(0)\|_2 \leq R$. Let $H(t,k)$ be defined by
\begin{align*}
    H(t,k)_{i,j} = H(t,k,c)_{i,j}
\end{align*}
for $j \in S_c$.
Then from Lemma~\ref{lem:gram-matrix-diff} we obtain 
\begin{align*}
    \|H(t,k)-H(0)\|_F \leq 2nR
\end{align*}
with probability at least $1-n^2\cdot \exp(-mR/10)$ over random initialization. 

Therefore from direct calculations we have
\begin{align*}
    &~\left| \sum_{i \in [n]}\sum_{k\in [K]}\sum_{c \in [N]}\sum_{j \in S_c}(y_i-y_i(t))(y_j-y_c^{(k)}(t)_j) (H(t,k,c)_{i,j}-H(0)_{i,j}) \right|\\
    = &~ \sum_{k \in [K]}(y-y(t))^\top (H(t,k)-H(0))(y-y^{(k)}(t))\\
    \leq&~ \sum_{k \in [K]}\|y-y(t)\|_2\|y-y^{(k)}(t)\|_2\|H(t,k)-H(0)\|_F\\
    \leq &~ 4nRK(1+2 \eta_{\loc} K n)\|y-y(t)\|_2^2.
\end{align*}
where the last step comes from Eq~\eqref{eq:bound-local-error}.

By Lemma~\ref{lem:bound-local-y-change} we have
\begin{align*}
    \left| \sum_{i \in [n]}\sum_{k\in [K]}\sum_{c \in [N]}\sum_{j \in S_c}(y_i-y_i(t))(y_j(t)-y_c^{(k)}(t)_j) H(0)_{i,j} \right| \leq&~ \sum_{k\in[K]}\|y-y(t)\|_2 \|H(0)\|\|y(t)-y^{(k)}(t)\|_2\\
    \leq&~ 2\eta_{\loc}  \kappa\lambda  K^2 n \|y-y(t)\|_2^2.
\end{align*}

Finally we have
\begin{align*}
    K\sum_{i \in [n]}\sum_{j \in [n]}(y_i-y_i(t))(y_j-y_j(t)) H(0)_{i,j} \geq &~ K\lambda \|y-y(t)\|_2^2.
\end{align*}
Combining the above we conclude the proof with
\begin{align*}
    & ~ C_1 \\
    =&~ -\frac{2\eta_{\glo}\eta_{\loc}}{N}\sum_{i \in [n]}\sum_{k\in [K]}\sum_{c \in [N]}\sum_{j \in S_c}(y_i-y_i(t))(y_j-y_c^{(k)}(t)_j) H(t,k,c)_{i,j} \\
    \leq &~-\frac{2\eta_{\glo}\eta_{\loc}}{N}(-4nRK(1+2 \eta_{\loc} K^2 n)\|y-y(t)\|_2^2 + K\lambda \|y-y(t)\|_2^2 - 2\eta_{\loc}  \kappa\lambda  K^2 n \|y-y(t)\|_2^2) \\
    \leq &~ \frac{2\eta_{\glo}\eta_{\loc}}{N}\|y-y(t)\|_2^2(-K\lambda + 4nRK(1+2 \eta_{\loc} K^2 n) +  2\eta_{\loc}  \kappa\lambda  K^2 n ).
    %\frac{2\eta_{\glo}}{N} \|y-y(t)\|_2^2 (-\lambda K \eta_{\loc} + \lambda \eta_{\loc}^2 K^2 n\kappa + 2\eta_{\loc} nR(1+\lambda \eta_{\loc} K^2 n\kappa)).
\end{align*}
\end{proof}

\begin{claim}\label{cla:bound_C2}
The following holds with probability at least $1-n\exp(-mR)$ over random initialization
\begin{align*}
    C_2 \leq \frac{16\eta_{\glo}\eta_{\loc}}{N}K(1+2\eta_{\loc} nK )nR\|y-y(t)\|_2^2.
\end{align*}
\iffalse
Further if 
\begin{align*}
    \eta_{\loc} \leq 1/(10 \lambda n K^2 )
\end{align*}
we have
\begin{align*}
C_2 \leq 40 \frac{\eta_{\glo} \eta_{\loc} K n R }{N}
\end{align*}
\fi
\end{claim}
\begin{proof}
We define matrix $H(t,k)^{\perp} \in \R^{n \times n}$ such that $H(t,k)^{\perp}_{i,j} = H(t,k,c)_{i,j}^{\perp}, j \in S_c$. Notice that
\begin{align*}
C_2 =&~ \frac{2\eta_{\glo}\eta_{\loc}}{N}\sum_{i \in [n]}\sum_{k\in [K]}\sum_{c \in [N]}\sum_{j \in S_c}(y_i-y_i(t))(y_j-y_c^{(k)}(t)_j) H(t,k,c)_{i,j}^{\perp}\\
=&~ \frac{2\eta_{\glo}\eta_{\loc}}{N}\sum_{k\in [K]}(y-y(t))^\top H(t,k)^\perp (y-y^{(k)}(t))\\
\leq &~ \frac{2\eta_{\glo}\eta_{\loc}}{N}\sum_{k\in [K]}\|y-y(t)\|_2 \|H(t,k)^\perp\|_F \|y-y^{(k)}(t)\|_2\\
\leq&~ \frac{4\eta_{\glo}\eta_{\loc}}{N}K(1+2\eta_{\loc} nK )\|y-y(t)\|_2^2 \|H(t,k)^{\perp}\|_F
\end{align*}
where the last step comes from Eq~\eqref{eq:bound-local-error}.

It thus suffices to upper bound $\| H(t,k)^{\bot} \|_F$.

For each $i \in [n]$, we define $\zeta_i$ as follows
\begin{align*}
\zeta_i=\sum_{r=1}^m\mathbf{1}_{r\in \ov{Q}_i} .
\end{align*}
It then follows from direct calculations that
\begin{align*}
\| H(t,k)^{\bot} \|_F^2
= & ~ \sum_{i=1}^n\sum_{j=1}^n (H(t,k)^{\bot}_{i,j})^2\\
= & ~ \sum_{i=1}^n\sum_{c \in [N]}\sum_{j \in S_c} \Big( \frac {1} {m}\sum_{r\in \ov{Q}_i} x_i^\top x_j\mathbf{1}_{u_r(t)^\top x_i\geq 0,w_{k,c,r}(t)^\top x_j\geq 0} \Big)^2\\
= & ~ \sum_{i=1}^n\sum_{c \in [N]}\sum_{j \in S_c} \Big( \frac {1} {m}\sum_{r=1}^m x_i^\top x_j\mathbf{1}_{u_r(t)^\top x_i\geq 0,w_{k,c,r}(t)^\top x_j\geq 0} \cdot \mathbf{1}_{r\in \ov{Q}_i} \Big)^2\\
= & ~ \sum_{i=1}^n\sum_{c \in [N]}\sum_{j \in S_c} ( \frac {x_i^\top x_j} {m} )^2 \Big( \sum_{r=1}^m \mathbf{1}_{u_r(t)^\top x_i\geq 0,w_{k,c,r}(t)^\top x_j\geq 0} \cdot \mathbf{1}_{r\in \ov{Q}_i} \Big)^2 \\
\leq & ~ \frac{1}{m^2} \sum_{i=1}^n\sum_{c \in [N]}\sum_{j \in S_c} \Big( \sum_{r=1}^m \mathbf{1}_{u_r(t)^\top x_i\geq 0,w_{k,c,r}(t)^\top x_j\geq 0} \cdot \mathbf{1}_{r\in \ov{Q}_i} \Big)^2 \\
= & ~ \frac{n}{m^2} \sum_{i=1}^n \Big( \sum_{r=1}^m \mathbf{1}_{r\in \ov{Q}_i} \Big)^2 \\
= & ~ \frac{n}{m^2} \sum_{i=1}^n \zeta_i^2 .
\end{align*}

Fix $i \in [n]$. The plan is to use Bernstein inequality to upper bound $\zeta_i$ with high probability.

First by Eq.~\eqref{eq:Air_bound} we have 
$
\E[\mathbf{1}_{r\in \ov{Q}_i}]\leq R 
$. 
We also have
\begin{align*}
\E \left[(\mathbf{1}_{r\in \ov{Q}_i}-\E[\mathbf{1}_{r\in \ov{Q}_i}])^2 \right]
 = & ~ \E[\mathbf{1}_{r\in \ov{Q}_i}^2]-\E[\mathbf{1}_{r\in \ov{Q}_i}]^2\\
\leq & ~ \E[\mathbf{1}_{r\in \ov{Q}_i}^2] \\
\leq & ~ R .
\end{align*}
Finally we have $|\mathbf{1}_{r\in \ov{Q}_i}-\E[\mathbf{1}_{r\in \ov{Q}_i}]|\leq 1$.

Notice that $\{\mathbf{1}_{r\in \ov{Q}_i}\}_{r=1}^m$ are mutually independent,
since $\mathbf{1}_{r\in \ov{Q}_i}$ only depends on $w_r(0)$.
Hence from Bernstein inequality (Lemma \ref{lem:bernstein}) we have for all $t>0$,
\begin{align*}
\Pr \left[ \zeta_i > m\cdot R+t \right] \leq \exp \left(-\frac{t^2/2}{m\cdot R+t/3} \right).
\end{align*}
By setting $t=3mR$, we have
\begin{align}\label{eq:Si_size_bound}
\Pr \left[ \zeta_i > 4mR \right] \leq \exp(-mR).
\end{align}
Hence by union bound,
with probability at least $1-n\exp(-mR)$,
\begin{align*}
\| H(t,k)^{\bot} \|_F^2 \leq \frac{n}{m^2}\cdot n\cdot (4mR)^2=16n^2R^2 .
\end{align*}
Putting all together we have
\begin{align*}
\| H(t,k)^{\bot} \|_F \leq 4nR
\end{align*}
with probability at least $1-n\exp(-mR)$ over random initialization.
\end{proof}

\begin{claim}\label{cla:bound_C3}
With probability at least $1-n\exp(-m R)$ over random initialization the following holds
\begin{align*}
    C_3 \leq \frac{16\eta_{\glo}\eta_{\loc}K}{N}(1+2\eta_{\loc} nK )nR\|y - y(t)\|_2^2
\end{align*}
\iffalse
Further if 
\begin{align*}
    \eta_{\loc} \leq 1/(10  n K^2 )
\end{align*}
we have
\begin{align*}
C_3 \leq 40 \frac{\eta_{\glo} \eta_{\loc} K n R }{N}\|y - y(t)\|_2^2
\end{align*}
\fi
\end{claim}
\begin{proof}
We can upper bound $\| v_2 \|_2$ in the following sense
\begin{align*}
\| v_2 \|_2^2
\leq &~ \sum_{i=1}^n \left(\frac{\eta_{\glo}}{ \sqrt{m} } \sum_{ r \in \ov{Q}_i } \left| \Delta u_r(t)^\top x_i \right|\right)^2\\
= &~ \frac{\eta_\glo^2}{ m }\sum_{i=1}^n \Big( \sum_{r=1}^m \mathbf{1}_{r\in \ov{Q}_i}| \Delta u_r(t)^\top x_i | \Big)^2\\
\leq &~ \frac{\eta_\glo^2\eta_\loc^2}{ m }\cdot \Big( \frac{2K(1+2\eta_\loc nK)\sqrt{n}}{N\sqrt{m}}\|y - y(t)\|_2 \Big)^2\cdot\sum_{i=1}^n \left(\sum_{r=1}^m \mathbf{1}_{r\in \ov{Q}_i}\right)^2
\end{align*}
where the last step comes from Lemma~\ref{lem:bound-local-y-change}.

It is previously shown that $\sum_{r=1}^m {\bf 1}_{r \in \ov{Q}_i } \leq 4 m R$ holds with probability at least $1-n\exp(-m R)$ over random initialization, thus with probability at least $1-n\exp(-m R)$ over random initialization
\begin{align*}
    \| v_2 \|_2^2 \leq&~ \frac{\eta_\glo^2\eta_\loc^2}{ m } \cdot \frac{4K^2(1+2\eta_{\loc} nK )^2n}{N^2m}\|y - y(t)\|_2^2 \cdot n( 4 m R)^2\\
    \leq &~ \Big( \frac{8\eta_{\glo}\eta_{\loc}K}{N}(1+2\eta_{\loc} nK )nR\|y - y(t)\| \Big)^2.
\end{align*}
Using Cauchy-Schwarz inequality, we complete the proof with
\begin{align*}
    C_3 =&~ - 2\sum_{i \in [n]}(y_i-y_i(t))v_{2,i}\\
    \leq &~ 2 \| y - y(t) \|_2 \cdot \| v_2 \|_2\\ \leq &~ \frac{16\eta_{\glo}\eta_{\loc}K}{N}(1+2\eta_{\loc} nK )nR\|y - y(t)\|_2^2.
\end{align*}
\end{proof}

\begin{claim}\label{cla:bound_C4}
We have
\begin{align*}
    C_4 \leq \frac{4\eta_{\loc}^2\eta_{\glo}^2n^2K^2(1+2\eta_\loc nK)^2}{N^2}\|y-y(t)\|^2_2.
\end{align*}
\end{claim}
\begin{proof}
Recall that $y(t+1)-y(t) = v_1 + v_2$, we have
\begin{align*}
\| y(t+1)-y(t) \|_2^2
\leq &~ \sum_{i=1}^n \left(\frac{\eta_{\glo}}{ \sqrt{m} } \sum_{ r =1 }^m \left| \Delta u_r(t)^\top x_i \right|\right)^2\\
= &~ \frac{\eta_\glo^2}{ m }\sum_{i=1}^n \Big( \sum_{r=1}^m | \Delta u_r(t)^\top x_i | \Big)^2\\
\leq &~ \frac{\eta_\glo^2\eta_\loc^2}{ m }\cdot \Big( \frac{2K(1+2\eta_\loc nK ) \sqrt{n}}{N\sqrt{m}}\|y - y(t)\|_2 \Big)^2\cdot nm^2\\
\leq &~ \frac{4\eta_{\loc}^2\eta_{\glo}^2n^2K^2(1+2\eta_\loc nK)^2}{N^2}\|y-y(t)\|^2_2
\end{align*}
where the penultimate step comes from Lemma~\ref{lem:bound-local-y-change}.
\end{proof}

\subsection{Random Initialization}
\begin{lemma}
Let events $E_1,E_2,E_3$ be defined as follows
\begin{align*}
    E_1 = &~ \Big\{ \phi(w_r(0)^\top x_i) \leq \sqrt{2\log (6mn/\delta)}, \forall r \in [m], \forall i \in [n] \Big\}\\
    E_2 = &~ \Big\{ \Big| \sum_{r = 1}^m \frac{1}{\sqrt{m}}a_r\phi(w_r(0)^\top x_i) {\bf 1}_{w_r(0)^\top x_i \leq \sqrt{2\log (6mn/\delta)}} \Big| \leq \sqrt{2\log ( 2mn / \delta) }\cdot \log ( 8n / \delta ), \forall i \in [n] \Big\}\\
    E_3 = &~ \Big\{ \sum_{r=1}^m {\bf 1}_{r \in \ov{Q}_i } \leq 4 m R,\forall i \in [n] \Big\}.
\end{align*}
Then $E_1\cap E_2\cap E_3$ is true with probability at least $1-\delta$ over the random initialization. Furthermore given $E_1\cap E_2\cap E_3$ the following holds
\begin{align*}
\|y-y(0)\|_2^2=O(n\log(m/\delta)\log^2(n/\delta)).
\end{align*}
\end{lemma}

\begin{proof}
First we bound $\Pr \left[ \neg E_3 \right]$. For each $i \in [n]$, we define $\zeta_i$ as follows
\begin{align*}
\zeta_i=\sum_{r=1}^m\mathbf{1}_{r\in \ov{Q}_i} .
\end{align*}

We use $w$ as shorthand for $w(0)$. Define the event
\begin{align*}
A_{i,r} = \left\{ \exists u : \| u - {w}_r \|_2 \leq R, {\bf 1}_{ x_i^\top {w}_r \geq 0 } \neq {\bf 1}_{ x_i^\top u \geq 0 } \right\}.
\end{align*}
Note this event happens if and only if $| {w}_r^\top x_i | < R$. Recall that ${w}_r \sim \N(0,I)$. By anti-concentration inequality of Gaussian (Lemma~\ref{lem:anti_gaussian}), we have
\begin{align}\label{eq:Air_bound}
\Pr[ A_{i,r} ] = \Pr_{ z \sim \N(0,1) } [ | z | < R ] \leq \frac{ 2 R }{ \sqrt{2\pi} }.
\end{align}

It thus follows from Eq.~\eqref{eq:Air_bound} that
$
\E[\mathbf{1}_{r\in \ov{Q}_i}]\leq R 
$. 
We also have
\begin{align*}
\E \left[(\mathbf{1}_{r\in \ov{Q}_i}-\E[\mathbf{1}_{r\in \ov{Q}_i}])^2 \right]
 = & ~ \E[\mathbf{1}_{r\in \ov{Q}_i}^2]-\E[\mathbf{1}_{r\in \ov{Q}_i}]^2\\
\leq & ~ \E[\mathbf{1}_{r\in \ov{Q}_i}^2] \\
\leq & ~ R .
\end{align*}
Therefore $|\mathbf{1}_{r\in \ov{Q}_i}-\E[\mathbf{1}_{r\in \ov{Q}_i}]|\leq 1$.

Notice that $\{\mathbf{1}_{r\in \ov{Q}_i}\}_{r=1}^m$ are mutually independent,
since $\mathbf{1}_{r\in \ov{Q}_i}$ only depends on $w_r$.
Hence from Bernstein inequality (Lemma \ref{lem:bernstein}) we have for all $t>0$,
\begin{align*}
\Pr \left[ \zeta_i > m\cdot R+t \right] \leq \exp \left(-\frac{t^2/2}{m\cdot R+t/3} \right).
\end{align*}
By setting $t=3mR$, we have
\begin{align}\label{eq:Si_size_bound_new}
\Pr \left[ \zeta_i > 4mR \right] \leq \exp(-mR).
\end{align}

Taking union bound and note the choice of $R$ and $m$ we have
\begin{align*}
    \Pr \left[ \neg E_3 \right] \leq n\exp(-mR) \leq \delta/3.
\end{align*}

Next we bound $\Pr[ \neg E_1]$. Fix $r\in [m]$ and $i\in [n]$.
Since $w_r\sim \N(0,I)$ and $\|x_i\|_2=1$,
$w_r^\top x_i$ follows distribution $\N(0,1)$.
From concentration of Gaussian distribution,
we have
\begin{align*}
\Pr_{w_r}[w_r^\top x_i\geq \sqrt{2\log (6mn / \delta) }]\leq \frac{\delta}{6mn}.
\end{align*}
Let $E_1$ be the event that
for all $r\in [m]$ and $i\in [n]$ we have
$
\phi(w_r^\top x_i)\leq \sqrt{2\log ( 6mn/ \delta) }.
$ 
Then by union bound,
$\Pr[ \neg E_1]\leq \frac {\delta}{3}$,

Finally we bound $\Pr[\neg E_2]$. Fix $i\in [n]$.
For every $r\in [m]$,
we define random variable $z_{i,r}$ as
\begin{align*}
z_{i,r}:=\frac {1}{\sqrt{m}} \cdot a_r \cdot \phi(w_r^\top x_i) \cdot \mathbf{1}_{w_r^\top x_i\leq \sqrt{2\log ( 6mn / \delta ) }}.
\end{align*}
Then $z_{i,r}$ only depends on $a_r\in \{-1,1\}$ and $w_r\sim \N(0,I)$.
Notice that $\E_{a_r,w_r}[z_{i,r}]=0$,
and $|z_{i,r}|\leq \sqrt{2\log ( 6mn / \delta ) }$.
Moreover,
\begin{align*}
 \E_{a_r,w_r}[z_{i,r}^2] 
= & ~\E_{a_r,w_r}\left[\frac {1}{m}a_r^2\phi^2(w_r^\top x_i)\mathbf{1}^2_{w_r^\top x_i\leq \sqrt{2\log ( 6mn / \delta) }}\right]\\
= & ~\frac {1}{m}\E_{a_r}[a_r^2] \cdot \E_{w_r} \Big[\phi^2(w_r^\top x_i)\mathbf{1}^2_{w_r^\top x_i\leq \sqrt{2\log ( 6mn / \delta) }} \Big]\\
\leq & ~\frac {1}{m}\cdot 1 \cdot \E_{w_r}[(w_r^\top x_i)^2] \\
= & ~ \frac {1} {m},
\end{align*}
where the second step uses independence between $a_r$ and $w_r$,
the third step uses $a_r\in \{-1,1\}$ and $\phi(t) = \max \{ t,0\}$,
and the last step follows from $w_r^\top x_i\sim \N(0,1)$.

Now we are ready to apply Bernstein inequality~(Lemma \ref{lem:bernstein}) to get for all $t>0$,
\begin{align*}
\Pr \left[ \sum_{r=1}^m z_{i,r}>t \right] \leq \exp\left(-\frac{t^2/2}{m\cdot \frac{1}{m}+\sqrt{2\log (6mn/\delta)} \cdot t/3} \right).
\end{align*}
Setting $t=\sqrt{2\log ( 6mn / \delta) }\cdot \log ( 8n / \delta )$,
we have with probability at least $1-\frac {\delta}{8n}$,
\begin{align*}
\sum_{r=1}^m z_{i,r}\leq \sqrt{2\log ( 6mn / \delta) }\cdot \log ( 8n / \delta ).
\end{align*}

Notice that we can also apply Bernstein inequality (Lemma~\ref{lem:bernstein}) on $-z_{i,r}$ to get
\begin{align*}
\Pr \left[ \sum_{r=1}^m z_{i,r}<-t \right] \leq \exp\left(-\frac{t^2/2}{m\cdot \frac{1}{m}+\sqrt{2\log (6mn/\delta)} \cdot t/3} \right).
\end{align*}
Let $E_2$ be the event that
for all $i\in[n]$,
\begin{align*}
\left| \sum_{r=1}^m z_{i,r} \right| \leq \sqrt{2\log ( 2mn / \delta) }\cdot \log ( 8n / \delta ).
\end{align*}
By applying union bound on all $i\in [n]$,
we have $\Pr[\neg E_2] \leq  \delta / 3$.

By union bound, $E_1\cap E_2\cap E_3$ will happen with probability at least $1-\delta$.

If both $E_1$ and $E_2$ happen,
we have
\begin{align*}
\|y-u(0)\|_2^2
= & ~ \sum_{i=1}^n(y_i-f(W(0),a,x_i))^2\\
= & ~ \sum_{i=1}^n \Big( y_i-\frac {1} {\sqrt{m}}\sum_{r=1}^{m} a_r\phi(w_r^\top x_i) \Big)^2\\
= & ~ \sum_{i=1}^n y_i^2-2\sum_{i=1}^n \frac{y_i}{\sqrt{m}}\sum_{r=1}^{m} a_r\phi(w_r^\top x_i) +\sum_{i=1}^n \frac {1}{m} \Big( \sum_{r=1}^{m} a_r\phi(w_r^\top x_i) \Big)^2\\
= & ~ \sum_{i=1}^n y_i^2-2\sum_{i=1}^n y_i\sum_{r=1}^{m} z_{i,r}+\sum_{i=1}^n \Big( \sum_{r=1}^{m}z_{i,r} \Big)^2\\
\leq & ~\sum_{i=1}^n y_i^2+2\sum_{i=1}^n |y_i|\sqrt{2\log ( 2mn / \delta) }\cdot \log ( 4n / \delta ) + \sum_{i=1}^n \Big( \sqrt{2\log ( 2mn / \delta) }\cdot \log ( 4n / \delta ) \Big)^2\\
= & ~ O(n\log(m/\delta)\log^2(n/\delta)) ,
\end{align*}
where the first step uses $E_1$, the second step uses $E_2$, and the last step follows from $|y_i| = O(1), \forall i \in [n]$.

\end{proof}
\subsection{Local Steps}
The following theorem is standard in neural tangent kernel theory (see e.g. \cite{sy19}).
\begin{lemma}\label{lem:fl-local}
With probability at least $1-\delta$ over the random initialization, the following holds for all $k \in [K]$ and $c \in [N]$ and $r\in [m]$ in step $t$
\begin{align}%\label{eq:fl-local}
\| y_c^{(k)} (t) - y_c \|_2^2 \leq &~ ( 1 - \eta_{\loc} \lambda / 2 )^k \cdot \| y_c^{(0)} (t) - y_c  \|_2^2,\\
\| w_{k,c,r}(t+1) - w_{0,c,r}(t) \|_2 \leq&~ \frac{ 4 \sqrt{n} \| y_c^{(0)} (t) - y_c  \|_2 }{ \sqrt{m} \lambda },\\
\| y_c^{(k+1)} (t) - y_c^{(k)} (t) \|_2^2 \leq&~ \eta_{\loc}^2 n^2 \cdot \| y_c^{(k)} (t) - y_c \|_2^2.\label{eq:local-y-move}
\end{align}
\end{lemma}

We then prove a Lemma that controls the updates in local steps.
\begin{lemma}\label{lem:bound-local-y-change}
Given Eq~\eqref{eq:local-y-move} for all $k \in [K], c \in [N]$ in step $t$ the following holds for all $k \in [K], c \in [N]$
\begin{align*}
    \|y_c(t)-y_c^{(k)}(t)\|_2 \leq &~ 2\eta_{\loc}nK \|y_c(t)-y_c\|,\\
    \|\Delta u_r(t)\|_2 \leq &~  \frac{2\eta_{\loc}K(1+2\eta_{\loc} nK )\sqrt{n}}{N\sqrt{m}}\|y - y(t)\|_2.
\end{align*}
\end{lemma}
\begin{proof}
For the first inequality, from Eq~\eqref{eq:local-y-move} we have
\begin{align*}
    \|y_c-y_c^{(k)}(t)\|_2 \leq&~ \|y_c^{(k)}(t)-y_c^{(k-1)}(t)\|_2 + \|y_c^{(k-1)}(t)-y_c\|_2\\
    \leq &~(\eta_{\loc}n + 1)\|y_c-y_c^{(k-1)}(t)\|_2\\
    \leq&~ (\eta_{\loc}n + 1)^k \|y_c-y_c(t)\|_2.
\end{align*}
Therefore 
\begin{align*}
    \|y_c(t)-y_c^{(k)}(t)\|_2 \leq &~ \sum_{i = 1}^k\|y_c^{(j)}(t)-y_c^{(j-1)}(t)\|_2\\
    \leq &~\sum_{i = 1}^k \eta_{\loc}n\|y_c-y_c^{(j-1)}(t)\|_2\\
    \leq &~ \sum_{i = 1}^k \eta_{\loc}n(\eta_{\loc}n + 1)^{j-1} \|y_c-y_c(t)\|_2\\
    \leq &~ 2\eta_{\loc}nK \|y_c(t)-y_c\|_2
\end{align*}
where the last step comes from the choice of $\eta_{\loc}$.

For the second inequality, notice that 
\begin{align*}
    \|\Delta u_r(t)\|_2 = &~ \eta_{\loc} \Big\|  \frac{a_r}{N}\sum_{c \in [N]}\sum_{k \in [K]} \frac{1}{\sqrt{m}}\sum_{j \in S_c}(y_j - y^{(k)}(t)_j)x_j{\bf 1} _{w_{r,k,c}(t)^\top x_j \geq 0} \Big\|_2 \\
    \leq &~ \frac{\eta_{\loc}}{N\sqrt{m}}\sum_{k \in [K]} \sum_{c \in [N]}\sum_{j \in S_c}|y_j - y^{(k)}(t)_j|\\
    \leq&~ \frac{\eta_{\loc}\sqrt{n}}{N\sqrt{m}}\sum_{k \in [K]} \|y - y^{(k)}(t)\|_2
\end{align*}
where the second step comes form triangle inequality and $\|x_i\|_2  = 1$ and the last step comes from Cauchy-Schwartz inequality.
From the $\|y_c(t)-y_c^{(k)}(t)\|_2 \leq 2\eta_{\loc}nK\|y_c(t)-y_c\|_2$ we have
\begin{align}\label{eq:bound-local-error}
    \notag \|y - y^{(k)}(t)\|_2^2  =&~ \sum_{c\in [N]}\|y_c - y_c^{(k)}(t)\|_2^2 = \sum_{c\in [N]}2(\|y_c - y_c(t)\|_2^2 + \|y_c(t)-y_c^{(k)}(t)\|_2^2)\\ \notag \leq&~ \sum_{c\in [N]}2(\|y_c - y_c(t)\|_2^2 + (2\eta_{\loc}nK)^2\|y_c(t)-y_c\|_2^2)\\ 
    \leq&~ 2(1+2\eta_{\loc} nK )^2\|y - y(t)\|_2^2.
\end{align}
It thus follows that
\begin{align*}
    \|\Delta u_r(t)\|_2 \leq&~ \frac{\eta_{\loc}\sqrt{n}}{N\sqrt{m}}\sum_{k \in [K]} \|y - y^{(k)}(t)\|_2\\
    \leq &~ \frac{2\eta_{\loc}K(1+2\eta_{\loc} nK )\sqrt{n}}{N\sqrt{m}}\|y - y(t)\|_2.
\end{align*}
\end{proof}

\subsection{Technical Lemma}\label{sec:appendix_tech}

\begin{lemma}
\label{lem:gram-matrix-diff}
For any set of weight vectors $\wt{w}_1, \cdots, \wt{w}_m \in \R^d$ and $\widehat{w}_1, \cdots, \widehat{w}_m \in \R^d$ define $H(\wt{w},\widehat{w}) \in \R^{n \times n}$ as 
\begin{align*}
    H(\wt{w},\widehat{w})_{i,j} =  \frac{1}{m} x_i^\top x_j \sum_{r=1}^m {\bf 1}_{ \wt{w}_r^\top x_i \geq 0, \widehat{w}_r^\top x_j \geq 0 } .
\end{align*}
Let $R \in (0,1)$ and ${w}_1, \cdots, {w}_m$ be iid generated from ${\N}(0,I)$. Then we have with probability at least $1-n^2 \cdot \exp(-m R /10)$ the following holds
\begin{align*}
\| H (w,w) - H(\wt{w},\widehat{w}) \|_F < 2 n R
\end{align*}
for any $\wt{w}_1, \cdots, \wt{w}_m \in \R^d$ and $\widehat{w}_1, \cdots, \widehat{w}_m \in \R^d$ such that $\| \widehat{w}_r - w_r \|_2 \leq R$ and $\| \wt{w}_r - w_r \|_2 \leq R$ for any $r\in [m]$.
\end{lemma}

\begin{proof}

For each $r \in [m]$ and $i,j \in [n]$, we define
\begin{align*}
s_{r,i,j} :=  {\bf 1}_{ \wt{w}_r^\top x_i \geq 0, \widehat{w}_r^\top x_j \geq 0} - {\bf 1}_{ w_r^\top x_i \geq 0 , w_r^\top x_j \geq 0 } .
\end{align*} 

The random variable we consider can be rewritten as follows
{%\footnotesize
\begin{align*}
& ~ \sum_{i=1}^n \sum_{j=1}^n | H(\wt{w},\widehat{w})_{i,j} - H(w,w)_{i,j} |^2 \\
\leq & ~ \frac{1}{m^2} \sum_{i=1}^n \sum_{j=1}^n \left( \sum_{r=1}^m {\bf 1}_{ \wt{w}_r^\top x_i \geq 0, \widehat{w}_r^\top x_j \geq 0} - {\bf 1}_{ w_r^\top x_i \geq 0 , w_r^\top x_j \geq 0 } \right)^2 \\
= & ~ \frac{1}{m^2} \sum_{i=1}^n \sum_{j=1}^n  \Big( \sum_{r=1}^m s_{r,i,j} \Big)^2.
\end{align*}
}
It thus suffices to bound $\frac{1}{m^2} ( \sum_{r=1}^m s_{r,i,j} )^2$.

Fix $i,j$ and we simplify $s_{r,i,j}$ to $s_r$.
Then $\{s_r\}_{r=1}^m$ are mutually independent random variables.

We define the event
\begin{align*}
A_{i,r} = \left\{ \exists u : \| u - {w}_r \|_2 \leq R, {\bf 1}_{ x_i^\top {w}_r \geq 0 } \neq {\bf 1}_{ x_i^\top u \geq 0 } \right\}.
\end{align*}

If $\neg A_{i,r}$ and $\neg A_{j,r}$ happen,
then 
\begin{align*}
\left| {\bf 1}_{ \wt{w}_r^\top x_i \geq 0, \widehat{w}_r^\top x_j \geq 0} - {\bf 1}_{ w_r^\top x_i \geq 0 , w_r^\top x_j \geq 0 } \right|=0.
\end{align*}
If   $A_{i,r}$ or $A_{j,r}$ happen,
then 
\begin{align*}
\left| {\bf 1}_{ \wt{w}_r^\top x_i \geq 0, \widehat{w}_r^\top x_j \geq 0} - {\bf 1}_{ w_r^\top x_i \geq 0 , w_r^\top x_j \geq 0 } \right|\leq 1.
\end{align*}
So we have {%\small
\begin{align*}
 \E_{{w}_r}[s_r]\leq \E_{{w}_r} \left[ {\bf 1}_{A_{i,r}\vee A_{j,r}} \right] 
 \leq & ~ \Pr[A_{i,r}]+\Pr[A_{j,r}] \\
 \leq & ~ \frac {4 R}{\sqrt{2\pi}} \\
 \leq & ~ 2 R,
\end{align*}}
and {%\small
\begin{align*}
    \E_{{w}_r} \left[ \left(s_r-\E_{{w}_r}[s_r] \right)^2 \right]
    = & ~ \E_{{w}_r}[s_r^2]-\E_{{w}_r}[s_r]^2 \\
    \leq & ~ \E_{{w}_r}[s_r^2]\\
    \leq & ~\E_{{w}_r} \left[ \left( {\bf 1}_{A_{i,r}\vee A_{j,r}} \right)^2 \right] \\
     \leq & ~ \frac {4R}{\sqrt{2\pi}} \\
     \leq  &~ 2 R .
\end{align*}}
We also have $|s_r|\leq 1$.
So we can apply Bernstein inequality (Lemma~\ref{lem:bernstein}) to get for all $t>0$,{%\small
\begin{align*}
    \Pr \left[\sum_{r=1}^m s_r\geq 2m R +mt \right]
    \leq & ~ \Pr \left[\sum_{r=1}^m (s_r-\E[s_r])\geq mt \right]\\
    \leq & ~ \exp \left( - \frac{ m^2t^2/2 }{ 2m R   + mt/3 } \right).
\end{align*}}
Choosing $t = R$, we get
\begin{align*}
    \Pr \left[\sum_{r=1}^m s_r\geq 3mR  \right]
    \leq & ~ \exp \left( -\frac{ m^2  R^2 /2 }{ 2 m  R + m  R /3 } \right) \\
     \leq & ~ \exp \left( - m R / 10 \right) .
\end{align*}
It follows that
\begin{align*}
\Pr \left[ \frac{1}{m} \sum_{r=1}^m s_r \geq 3  R \right] \leq \exp( - m R /10 ).
\end{align*}
Similarly
\begin{align*}
\Pr \left[ \frac{1}{m} \sum_{r=1}^m s_r \leq -3 R \right] \leq \exp( - m R /10 ).
\end{align*}
Therefore we complete the proof.
\end{proof}

\begin{lemma}\label{lem:bound-global-weights}
If Eq.~\eqref{eq:fl-global} holds for $i = 0, \cdots, k$, then we have for all $r\in [m]$
\begin{align*}
\| u_r(t) - u_r(0) \|_2 \leq \frac{ 8 \sqrt{n} \| y - y (0) \|_2 }{ \sqrt{m} \lambda } := D.
\end{align*}
\end{lemma}
\begin{proof}
We have 
\begin{align*}
    \| u_r(t) - u_r(0) \|_2 \leq &~ \eta_{\glo}\sum_{\tau = 0}^t \|\Delta u_r(\tau)\|_2\\
    \leq &~ \eta_{\glo}\sum_{\tau = 0}^t \frac{2\eta_{\loc}K(1+2\eta_{\loc} nK )\sqrt{n}}{N\sqrt{m}}\|y - y(\tau)\|_2\\
    \leq &~ \eta_{\glo} \frac{2\eta_{\loc}K(1+2\eta_{\loc} nK )\sqrt{n}}{N\sqrt{m}}\sum_{\tau = 0}^t (1-\frac{\eta_{\glo}\eta_{\loc}\lambda K}{2N})^{\tau}\|y - y(0)\|_2\\
    \leq&~ \frac{ 8 \sqrt{n} \| y - y (0) \|_2 }{ \sqrt{m} \lambda }.
\end{align*}
where the second step comes from Lemma~\ref{lem:bound-local-y-change} and the last step comes from the choice of $\eta_{\loc}$.
\end{proof}

%!TEX root=main.tex
\section{Generalization}\label{sec:generalization}

In this section, we generalize our initialization scheme to each $w_r(0) \sim {\N}(0,\sigma^2 I)$. Notice that this just introduces an extra $\sigma^{-2}$ term to every occurrence of $m$. %Similar to Appendix~\ref{sec:full_proof}, it is not hard to obtain the following lemma. 
In addition, we use $U(t) = [u_1(t), \cdots, u_m(t)]^\top \in \R^{d \times m}$ to denote parameters in a matrix form.
For convenience, we first list several definitions and results which will be used in the proof our generalization theorem. Our setting mainly follows \cite{adh+19a}.

\subsection{Definitions}

\begin{definition}[Non-degenerate Data Distribution, Definition 5.1 in \cite{adh+19a}]
A distribution $\mathcal{D}$ over $\mathbb{R}^d\times\mathbb{R}$ is \emph{$(\lambda,\delta,n)$-non-degenerate},
if with probability at least $1-\delta$,
for $n$ iid samples $\{ (x_i,y_i) \}_{i=1}^n$ chosen from $\mathcal{D}$,
$\lambda_{\min}(H^{\infty})\geq \lambda>0$.
\end{definition}

%We adapt the notations on the loss functions from \cite{adhlw19}.
\begin{definition}[Loss Functions]
Let $\ell:\mathbb{R}\times \mathbb{R}\rightarrow \mathbb{R}$ be the loss function.
For function $f:\mathbb{R}^d\rightarrow \mathbb{R}$,
for distribution $\mathcal{D}$ over $\mathbb{R}^d\times\mathbb{R}$,
the \emph{population loss} is defined as
\begin{align*}
L_{\mathcal{D}}(f):=\E_{(x,y)\sim \mathcal{D}}[\ell(f(x),y)].
\end{align*}
Let $S=\{(x_i,y_i)\}_{i=1}^n$ be $n$ samples.
The empirical loss over $S$ is defined as
\begin{align*}
L_S(f):=\frac{1}{n}\sum_{i=1}^n \ell(f(x_i),y_i).
\end{align*}
\end{definition}

%Rademacher complexity is a useful tool to work with the generalization error. Here we give the definition.
\begin{definition}[Rademacher Complexity]
Let $\mathcal{F}$ be a class of functions mapping from $\mathbb{R}^d$ to $\mathbb{R}$.
Given $n$ samples $S=\{x_1,\cdots,x_n\}$ where $x_i\in \mathbb{R}^d$ for $i\in [n]$,
the \emph{empirical Rademacher complexity} of $\mathcal{F}$ is defined as
\begin{align*}
\mathcal{R}_S(\mathcal{F}):=\frac{1}{n}\E_{\epsilon}\left[\sup_{f\in \mathcal{F}}\sum_{i=1}^n \epsilon_i f(x_i)\right].
\end{align*}
where $\epsilon\in \mathbb{R}^d$ and each entry of $\epsilon$ are drawn from independently uniform at random from $\{\pm 1\}$.
\end{definition}

\subsection{Tools from Previous Work}

\begin{theorem}[Theorem B.1 in \cite{adh+19a}]\label{thm:sample_to_generalization}
Suppose the loss function $\ell(\cdot,\cdot)$ is bounded in $[0,c]$ for some $c>0$ and is $\rho$-Lipschitz in its first argument.
Then with probability at least $1-\delta$ over samples $S$ of size $n$,
\begin{align*}
\sup_{f\in \mathcal{F}} \{L_{\mathcal{D}}(f)-L_S(f)\}\leq 2\rho \mathcal{R}_S(\mathcal{F})+3c\sqrt{\frac{\log(2/\delta)}{2n}}.
\end{align*}
\end{theorem}

\begin{lemma}[Lemma 5.4 in \cite{adh+19a}]\label{lem:rademacher_upper_bound}
Given $R>0$,
with probability at least $1-\delta$ over the random initialization on $U(0)\in \mathbb{R}^{m\times d}$ and $a\in \mathbb{R}^m$,
for all $B>0$,
the function class
\begin{align*}
\mathcal{F}_{R,B}^{U(0),a}=\{f(U,\cdot,a) = \frac{1}{ \sqrt{m} } \sum_{r=1}^m a_r \phi ( u_r^\top x ):\|u_r-u_r(0)\|_2\leq R, \forall r\in [m]; \|U-U(0)\|_F\leq B\}
\end{align*}
has bounded empirical Rademacher complexity
\begin{align*}
\mathcal{R}_S(\mathcal{F}_{R,B}^{U(0),a})\leq \frac{B}{\sqrt{2n}}\left(1 + \Big( \frac{2\log(2/\delta)}{m} \Big )^{1/4}\right)+\frac{2R^2\sqrt{m}}{\sigma}+R\sqrt{2\log(2/\delta)}.
\end{align*}
\end{lemma}

\begin{lemma}[Lemma C.3 in \cite{adh+19a}]\label{lem:initial-H-concentration}
With probability at least $1-\delta$ we have
\begin{align*}
    \|H(0) - H^{\infty}\|_F \leq O(n\sqrt{\log(n/\delta)/\sqrt{m}}).
\end{align*}

\end{lemma}
\subsection{Complexity Bound}

To simplify the proof in the following sections, we define $\rho:= {\eta_\loc \eta_\glo K} / {N}$.

Now we prove a key technical lemma which will be used to prove the main result.
\begin{lemma}%[Improved version of Lemma 5.3 in \cite{adh+19a}]
\label{lem:total_movement}
Let $\lambda =\lambda_{\min}(H^{\infty})>0$.
Fix $\sigma>0$, let $m = \Omega( \lambda^{-4} \sigma^{-2} n^4 \log (n/\delta) )$, we iid initialize $w_r \sim  {\N}(0,\sigma^2 I)$, $a_r$ sampled from $\{-1,+1\}$ uniformly at random for $r\in [m]$ and set $\eta_\loc =O(\frac{\lambda}{n^2 K \kappa}),\eta_\glo = O(1)$.
For weights $w_1,\cdots,w_m\in \mathbb{R}^d$, let $\vect(W)=[w_1^\top \, w_2^\top \,\cdots w_m^\top]^\top\in \mathbb{R}^{md}$ be the concatenation of $w_1,\cdots,w_m$.
Then with probability at least $1-6 \delta$ over the random initialization,
we have for all $t \geq 0$,
\begin{itemize}
\item $\| u_r(t) - u_r(0) \|_2 \leq \frac{ 8 \sqrt{n} \| y - u (0) \|_2 }{ \sqrt{m} \lambda }$,
\item $\|U(t)-U(0)\|_F\leq (y^\top (H^{\infty})^{-1} y)^{1/2}+O\left( ( \frac{n\sigma }{\lambda} +\frac{n^{7/2}}{\sigma^{1/2} {m}^{1/4}}) \cdot \poly( \log(m/\delta))\right)$.
\end{itemize}
\end{lemma}
\begin{proof}
Similarly to Appendix~\ref{sec:full_proof}, $\| u_r(t) - u_r(0) \|_2 \leq \frac{ 8 \sqrt{n} \| y - u (0) \|_2 }{ \sqrt{m} \lambda }$ and $\|w_{k,c,r}(t)-u(0)\|_2 \leq \frac{8\sqrt{n} \| y - u (0) \|_2 }{ \sqrt{m} \lambda }$.
For integer $k\geq 0$,
define $J(k,t)\in \mathbb{R}^{md\times n}$ as the matrix
\begin{align*}
J(k,t)=\frac{1}{\sqrt{m}} \begin{pmatrix} a_1x_1 {\bf 1}_{ w_{k,c_1,1}(t)^\top x_1 \geq 0} & \cdots & a_1x_n {\bf 1}_{ w_{k,c_n,1}(t)^\top x_n \geq 0}\\
\vdots & \ddots & \vdots\\
a_m x_1 {\bf 1}_{ w_{k,c_1,m}(t)^\top x_1 \geq 0} & \cdots & a_mx_n {\bf 1}_{ w_{k,c_n,m}(t)^\top x_n \geq 0}\\
\end{pmatrix}
\end{align*}
where $c_i \in [N]$ denotes the unique client such that $i \in c_i$.
We claim that 
\begin{align*}
\|J(k,t) - J(0,0)\|_F \leq O\Big(n\cdot \big(\delta+ \frac{n\sqrt{\log(m/\delta)\log^2(n/\delta)}}{\sigma \lambda \sqrt{m}}\big) \Big)^{1/2}.
\end{align*}
In fact, we can calculate $\|J(k,t) - J(0,0)\|_F^2$ in the following
\begin{align*}
    \|J(k,t) - J(0,0)\|_F^2 = &~ \frac{1}{m}\sum_{r = 1}^{m}\sum_{c \in [N]}\sum_{i \in S_c}\left(\|x_i\|_2 \cdot a_i (\mathbf{1}_{w_{k,c,r}(t)^\top x_i \geq 0} - \mathbf{1}_{u_r(0)^\top x_i \geq 0})\right)^2\\
    = &~ \frac{1}{m}\sum_{r = 1}^{m}\sum_{c \in [N]}\sum_{i \in S_c}\left(\mathbf{1}_{w_{k,c,r}(t)^\top x_i \geq 0} - \mathbf{1}_{u_r(0)^\top x_i \geq 0}\right)^2\\
    = &~ \frac{1}{m}\sum_{r = 1}^{m}\sum_{c \in [N]}\sum_{i \in S_c} \mathbf{1}_{\mathbf{1}_{w_{k,c,r}(t)^\top x_i \geq 0} \neq \mathbf{1}_{u_r(0)^\top x_i \geq 0}}.
\end{align*}
Fix ${c \in [N]}, {i \in S_c}$ and for $r \in [m]$ define $t_r$ as follows
\begin{align*}
t_r = \mathbf{1}_{\mathbf{1}_{w_{k,c,r}(t)^\top x_i \geq 0} \neq \mathbf{1}_{u_r(0)^\top x_i \geq 0}}.
\end{align*}

Consider the event
\begin{align*}
    A_{i,r} = \{\exists w: ~ \|u_r(0) - w\|_2 \leq R, {\mathbf{1}_{w^\top x_i \geq 0} \neq \mathbf{1}_{u_r(0)^\top x_i \geq 0}} \}
\end{align*}
where $R = \frac{C n\sqrt{\log(m/\delta)\log^2(n/\delta)}}{\lambda \sqrt{m}}$ for sufficiently small constant $C >0$.
If $t_r = 1$ then either $A_{i,r}$ happens or $\|w_{k,c,r}(t)-u_r(0)\|_2 \leq R$, otherwise $t_r = 0$. Therefore
\begin{align*}
    \E[t_r] \leq \Pr[A_{i,r}] + \Pr[\|u_r(0) - w_{k,c,r}(t)\|_2 < R] \leq R\sigma^{-1}  + \delta.
\end{align*}
And similarly $\E[(t_r - \E[t_r])^2] \leq \E[t_r^2] = R\sigma^{-1}  + \delta$.
Applying Bernstein inequality, we have for all $t >0$,
\begin{align*}
    \Pr \Big[ \sum_{r=1}^m t_r \geq mR\sigma^{-1} + m\delta + mt \Big] \leq \exp \left(-\frac{m^2t^2}{2(mR\sigma^{-1} + m\delta + mt/3)}\right).
\end{align*}
Choosing $t = R\sigma^{-1} +\delta$,
\begin{align*}
    \Pr \Big[ \sum_{r=1}^m t_r \geq 2m(R\sigma^{-1} + \delta) \Big] \leq \exp (-m(R\sigma^{-1}+ \delta)/10).
\end{align*}
By applying union bound over $i \in [n]$, we have with probability at least $1-n\exp (-m(R\sigma^{-1}+ \delta)/10)$, $\|J(k,t) - J(0,0)\|_F \leq 2n(R\sigma^{-1} + \delta)$. This is exactly what we need.

Notice that we can rewrite the update rule in federated learning as
\begin{align}\label{eq:vec_grad_update}
\vect(U(t+1))= & ~ \vect(U(t))-\frac{\eta_\glo}{N}\sum_{k \in [K]}\eta_\loc  J(k,c)(y^{(k)}(t)-y) \notag \\
= & ~  \vect(U(t))- \rho \frac{1}{K} \sum_{k \in [K]}  J(k,c)(y^{(k)}(t)-y)
\end{align}
where the last step follows from definition of $\rho = \eta_{\glo} \eta_{\loc} K / N$.

Recall from Appendix~\ref{sec:full_proof} that 
\begin{align*}
    \Delta u_r(t) = \frac{a_r}{N}\sum_{c \in [N]}\sum_{k \in [K]} \frac{\eta_{\loc}}{\sqrt{m}}\sum_{j \in S_c}(y_j - y_c^{(k)}(t)_j)x_j{\bf 1} _{w_{k,c,r}(t)^\top x_j \geq 0},
\end{align*}
and
\begin{align*}
     H(t,k,c)_{i,j} =&~ \frac{1}{m} \sum_{r=1}^{m} x_{i}^\top x_j {\bf 1} _{u_{r}^\top x_i \geq 0,w_{k,c,r}(t)^\top x_j \geq 0},\\
     H(t,k,c)_{i,j}^{\perp} =&~ \frac{1}{m} \sum_{r \in \ov{Q}_i} x_{i}^\top x_j {\bf 1}_{u_{r}^\top x_i \geq 0,w_{k,c,r}(t)^\top x_j \geq 0}.
\end{align*}
We have
\begin{align*}
    v_{1,i} = &~ \frac{1}{\sqrt{m}}\sum_{r \in Q_i}a_r \bigg(\phi\big((u_r(t)+\eta_{\glo}\Delta u_r(t))^\top x_i\big) - \phi(u_r(t)^\top x_i)\bigg)\\
    = &~ -\frac{\eta_\loc \eta_\glo K}{N}\sum_{j \in S_c} (y(t)_j-y_j)H^{\infty}_{i,j} + (-\frac{\eta_\loc \eta_\glo }{N})\sum_{k \in [K]}\sum_{c \in [N]}\sum_{j \in S_c} (y^{(k)}(t)_j-y(t)_j)H^{\infty}_{i,j}\\
    &~ + (-\frac{\eta_\loc \eta_\glo }{N}) \sum_{k \in [K]}\sum_{c \in [N]}\sum_{j \in S_c} (y^{(k)}(t)_j-y_j)(H(t,k,c)_{i,j}-H^{\infty}_{i,j})\\
    &~ + (-\frac{\eta_\loc \eta_\glo }{N}) \sum_{k \in [K]}\sum_{c \in [N]}\sum_{j \in S_c} (y^{(k)}(t)_j-y_j)H(t,k,c)_{i,j}^{\perp},\\
    v_{2,i} = &~ \frac{1}{\sqrt{m}}\sum_{r \in \ov{Q}_i}a_r \bigg(\phi\big((u_r(t)+\eta_{\glo}\Delta u_r(t))^\top x_i\big) - \phi(u_r(t)^\top x_i)\bigg).
\end{align*}

Following the proof of Appendix~\ref{sec:full_proof}, let
\begin{align*}
    \xi_i(t) = &~ v_{2,i}(t) + (-\frac{\eta_\loc \eta_\glo }{N})\sum_{k \in [K]}\sum_{c \in [N]}\sum_{j \in S_c} (y^{(k)}(t)_j-y(t)_j)H^{\infty}_{i,j}\\
    &~ + (-\frac{\eta_\loc \eta_\glo }{N}) \sum_{k \in [K]}\sum_{c \in [N]}\sum_{j \in S_c} (y^{(k)}(t)_j-y_j)(H(t,k,c)_{i,j}-H^{\infty}_{i,j})\\
    &~ + (-\frac{\eta_\loc \eta_\glo }{N}) \sum_{k \in [K]}\sum_{c \in [N]}\sum_{j \in S_c} (y^{(k)}(t)_j-y_j)H(t,k,c)_{i,j}^{\perp}.
\end{align*}
Notice that
\begin{align*}
\sum_{i=1}^n |\ov{Q}_i|=\sum_{r=1}^m \sum_{i=1}^n{\bf 1}_{r\in \ov{Q}_i}=\sum_{i=1}^n(\sum_{r=1}^m {\bf 1}_{r\in \ov{Q}_i}).
\end{align*}
Hence by Eq. \eqref{eq:Si_size_bound},
with probability at least $1-n\exp(-mR\sigma^{-1})$ we have
\begin{align*}
\sum_{i=1}^n |\ov{Q}_i|\leq 4mnR\sigma^{-1}.
\end{align*}
Similar to Appendix~\ref{sec:full_proof}, by the choice of $R=\frac{ 8 \sqrt{n} \| y - u (0) \|_2 }{ \sqrt{m} \lambda }$ and 
\begin{align*}
\|y-y(0)\|_2=O \Big( \sqrt{n\log(m/\delta)\log^2(n/\delta)} \Big), 
\end{align*}
we can bound $\xi(t) = [\xi_1(t),\cdots,\xi_n]^\top \in \R^n$ as
\begin{align}\label{eq:bound-xi}
    \|\xi(t)\|_2 \leq & ~ O\bigg(\frac{\eta_\glo \eta_\loc n^{5/2}K\kappa \sqrt{\log(m/\delta)\log^2(n/\delta)}}{N\sigma\lambda \sqrt{m}}\|y-y(t)\|_2\bigg) \notag \\
    = & ~ O\bigg(\frac {\rho n^{5/2}\kappa \sqrt{\log(m/\delta)\log^2(n/\delta)}}{\sigma\lambda \sqrt{m}}\|y-y(t)\|_2\bigg)
\end{align}
where the last step follows from definition of $\rho = \eta_{\glo} \eta_{\loc} K / N$.

Notice that with probability at least $1-\delta$, for all $i\in[n]$,
\begin{align*}
|y_i(0)|\leq \sigma \cdot \sqrt{2\log ( 2mn / \delta) }\cdot \log ( 4n / \delta ),
\end{align*}
which implies
\begin{align}\label{eq:bound-y0}
\|y(0)\|_2^2 \leq n\sigma^2 \cdot 2\log ( 2mn / \delta) \cdot \log^2 ( 4n / \delta ).
\end{align}

Therefore we can explicitly write the dynamics of the global model as
\begin{align*}
    y(t)-y = &~ (I - \frac{\eta_\glo\eta_\loc K}{N} H^\infty )(y(t-1)-y) + \xi(t-1)\\
    = &~ (I - \rho H^\infty )(y(t-1)-y) + \xi(t-1)\\
    = &~ (I- \rho  H^\infty )^t (y(0)-y) + \sum_{\tau = 0}^{t-1}(I- \rho H^\infty )^{\tau} \xi(t-1-\tau)\\
    =&~ -(I- \rho H^\infty )^t y + e(t).
\end{align*}
where the second step follows from definition of $\rho = \eta_{\glo} \eta_{\loc} K / N$, the third step comes from recursively applying the former step.y.

By Eq~\eqref{eq:bound-xi} and Eq~\eqref{eq:bound-y0} we have
\begin{align*}
    e(t) = &~ (I- \rho H^\infty )^t y(0) + \sum_{\tau = 0}^{t-1}(I- \rho H^\infty )^{\tau} \xi(t-1-\tau)\\
    = &~ O\bigg( (1- \rho )^t\cdot \sqrt{n\sigma^2}\cdot \sqrt{2\log(2mn/\delta)} \cdot \log(8n/\delta) + t(1- \rho )^t\cdot\frac{ \rho n^3 \log(m/\delta)\log^2(n/\delta)}{\lambda \sigma \sqrt{m}}\bigg)
\end{align*}
where we used $\| y (t) - y \|_2^2 \leq ( 1 - \frac{\eta_{\glo} \eta_{\loc} \lambda K}{2N} )^t \cdot \| y (0) - y \|_2^2$ from Theorem~\ref{thm:fl-ntk-convergence}.

By Eq~\eqref{eq:vec_grad_update},
\begin{align*}
\vect(U(T))-\vect(U(0))= & ~\sum_{t=0}^{T-1}(\vect(U(t+1))-\vect(U(t)))\\
= & ~\sum_{t=0}^{T-1}\left(- \rho \cdot \frac{1}{K} \sum_{k \in [K]} J(k,t)(y^{(k)}(t)-y)\right)\\
= & ~ \sum_{t=0}^{T-1} \rho \cdot J(0,0)(I- \rho  H^\infty )^t y\\
& ~+ \sum_{t=0}^{T-1} \rho \cdot \frac{1}{K}\sum_{k \in [K]} (J(k,t)-J(0,0))(I- \rho H^\infty )^t y\\
&~ - \sum_{t=0}^{T-1} \rho \cdot \frac{1}{K} \sum_{k \in [K]} J(k,t) (y^{(k)}(t)-y(t) + e(k))\\
= & ~B_1+B_2+B_3
\end{align*}
where 
\begin{align*}
B_1:= & ~ + \rho \cdot \sum_{t=0}^{T-1}   J(0,0)(I- \rho H^\infty )^t y,\\
B_2:= & ~ + \rho \cdot \sum_{t=0}^{T-1} \frac{1}{K} \sum_{k \in [K]} (J(k,t)-J(0,0))(I- \rho H^\infty )^t y,\\
B_3:= & ~ - \rho \cdot \sum_{t=0}^{T-1}  \frac{1}{K} \sum_{k \in [K]} J(k,t)(y^{(k)}(t)-y(t) + e(k)). %\\
%\rho:= & ~ {\eta_\loc \eta_\glo K }/{N}
\end{align*}
We bound these terms separately.

Putting Claim~\ref{cla:B_1}, \ref{cla:B_2} and \ref{cla:B_3} together
we have
\begin{align*}
& ~\|U(T)-U(0)\|_F \\
= & ~\|\vect(U(T))-\vect(U(0))\|_2\\
= & ~ B_1 + B_2 + B_3 \\
\leq & ~(y^\top (H^{\infty})^{-1} y)^{1/2}+O\left((\frac{n^2\sqrt{\log(n/\delta)}}{\lambda^2 \sqrt{m}})^{1/2} + (\frac{n^{3/2}}{m^{1/4}\sigma^{1/2}\lambda^{3/2}} + \frac{n\sigma }{\lambda}+\frac{n^{7/2}}{\lambda^3 \sigma \sqrt{m}}) \cdot \poly(\log(m/\delta))\right)\\
\leq & ~(y^\top (H^{\infty})^{-1} y)^{1/2}+O\left( \frac{n\sigma }{\lambda}\cdot \poly(\log(m/\delta))+\frac{n^{7/2}}{\sigma^{1/2} {m}^{1/4}} \cdot \poly(\log(m/\delta))\right)
\end{align*}
which completes the proof of Lemma \ref{lem:total_movement}.
\end{proof}

\subsection{Technical Claims}

\begin{claim}[Bounding $B_1$]\label{cla:B_1}
With probability at least $1-\delta$ over the random initialization,
we have
\begin{align*}\label{eq:B_1}
\|B_1\|_2^2 \leq y^\top D^\top H^{\infty} D y+O(\frac{n^2\sqrt{\log(n/\delta)}}{\lambda^2 \sqrt{m}})
\end{align*}
where $D=\sum_{t=0}^{T-1} \rho (I- \rho H^{\infty})^t \in \mathbb{R}^{n\times n}$.
\end{claim}
\begin{proof}
Recall $D=\sum_{t=0}^{T-1} \rho (I- \rho H^{\infty})^t \in \mathbb{R}^{n\times n}$,
then we have
\begin{align*}%\label{eq:B_1}
\|B_1\|_2^2= & ~ \Big\| \sum_{t=0}^{T-1} \rho \cdot J(0,0) \cdot (I- \rho \cdot H^{\infty})^t y \Big\|_2^2 \notag \\
= & ~y^\top D^\top J(0,0)^\top J(0,0) Dy \notag \\
= & ~y^\top D^\top H^{\infty} Dy+y^\top D^\top (H(0)-H^{\infty}) Dy \notag \\
\leq &~ y^\top D^\top H^{\infty} Dy+ \|H(0)-H^{\infty}\|_F \cdot \|D\|_2^2 \|y\|^2 \notag \\
\leq & ~ y^\top D^\top H^{\infty} D y+O(\frac{n\sqrt{\log(n/\delta)}}{ \sqrt{m}}) \left(\sum_{t=0}^{T-1} \rho (1- \rho \lambda)^t\right)^2 n\\
\leq &~ y^\top D^\top H^{\infty} D y+O(\frac{n^2\sqrt{\log(n/\delta)}}{\lambda^2 \sqrt{m}})
\end{align*}
where the penultimate step comes from Lemma~\ref{lem:initial-H-concentration} and $y_i = O(1)$.
\end{proof}

\begin{claim}[Bounding $B_2$]\label{cla:B_2}
With probability at least $1-\delta$ over the random initialization,
we have
\begin{align*}
\|B_2\|_2 \leq  \frac{n^{3/2}\poly(\log(m/\delta))}{m^{1/4}\sigma^{1/2}\lambda^{3/2}} .
\end{align*}

\end{claim}

\begin{proof}
For $B_2$,
we have
\begin{align*}
\|B_2\|_2= & ~ \Big\|\sum_{t=0}^{T-1} \rho \cdot \frac{1}{K} \sum_{k \in [K]} (J(k,t)-J(0,0))(I- \rho H^\infty )^t y \Big\|_2 \notag \\
\leq & ~ \sum_{t=0}^{T-1} \rho \cdot \frac{1}{K} \sum_{k \in [K]} \|J(k,t)-J(0,0)\|_F \cdot \|I - \rho H^\infty \|_2^t \cdot \|y\|_2 \notag\\
\leq & ~ O\left(\frac{n\poly(\log(m/\delta))}{m^{1/4}\sigma^{1/2}\lambda^{1/2}} \cdot \rho \cdot 
\sum_{k=0}^{K-1} (1- \rho \lambda)^k\cdot \sqrt{n}\right) \notag\\
= & ~O\left(\frac{n^{3/2}\poly(\log(m/\delta))}{m^{1/4}\sigma^{1/2}\lambda^{3/2}}\right).
\end{align*}
where in the third step we use
\begin{align*}
\|J(k,t) - J(0,0)\|_F \leq O\Big(n\cdot \big(\delta+ \frac{n\sqrt{\log(m/\delta)\log^2(n/\delta)}}{\sigma \lambda \sqrt{m}} \big) \Big)^{1/2}
\end{align*}
and without loss of generality, we can set $\delta$ sufficiently small.
\end{proof}

\begin{claim}[Bounding $B_3$]\label{cla:B_3}
With probability at least $1-\delta$ over the random initialization,
we have
\begin{align*}%\label{cla:eq:B_3}
\|B_3\|_2 \leq  (\frac{n\sigma }{\lambda}+\frac{n^{7/2}}{\lambda^3 \sigma \sqrt{m}}) \cdot \poly(\log(m/\delta)).
\end{align*}
\end{claim}
\begin{proof}
Notice that for $k,t\geq 0$,
$\|J(k,t)\|_F^2 \leq \frac{mn}{m}=n$. By Eq~\eqref{eq:bound-xi} and Eq~\eqref{eq:bound-y0} we have
\begin{align*}%\label{eq:B_3}
\|B_3\|_2= & ~\left\|-\sum_{t=0}^{T-1} \rho \cdot \frac{1}{K} \sum_{k \in [K]} J(k,t)(y^{(k)}(t)-y(t) + e(k))\right\|_2\notag \\
\leq & ~ \rho \frac{1}{K} \cdot \sqrt{n} \cdot \sum_{t=0}^{T-1} O\bigg( (1-\rho)^t\cdot \sqrt{n\sigma^2}\cdot \sqrt{2\log(2mn/\delta)} \cdot \log(8n/\delta) \notag \\
&~ + t(1-\rho)^t\cdot\frac{ \rho n^3 \log(m/\delta)\log^2(n/\delta)}{\lambda \sigma \sqrt{m}} \bigg) \notag \\
\leq & ~  (\frac{n\sigma }{\lambda}+\frac{n^{7/2}}{\lambda^3 \sigma \sqrt{m}}) \cdot \poly(\log(m/\delta)),
\end{align*}
here in the first step $-\sum_{t=0}^{T-1} \rho \cdot \frac{1}{K} \sum_{k \in [K]} J(k,t) e(k)$ is the dominant term.

\end{proof}

\subsection{Main Results}
Now we can present our main result in this section.
\begin{theorem}\label{thm:fl-ntk-generalization}
Fix failure probability $\delta \in (0,1)$. Set  $\sigma=O( \lambda \poly(\log n,\log(1/\delta))  / {n})$,
$m = \Omega\left( \sigma^{-2}(n^{14}\poly(\log m,\log(1/\delta),\lambda^{-1})) \right)$, let the two layer neural network be initialized with $w_r$ i.i.d sampled from ${\N}(0,\sigma^2 I)$ and $a_r$ sampled from $\{-1,+1\}$ uniformly at random for $r\in [m]$. 
Suppose the training data $S=\{(x_i,y_i)\}_{i=1}^n$ are i.i.d samples from a $(\lambda,\delta/3,n)$-non-degenerate distribution $\mathcal{D}$. Let $\rho= \eta_{\loc} \eta_{\glo} K / N$ and train the two layer neural network $f(U(t),\cdot,a)$ by federated learning for
\begin{align*}
T \geq \Omega\left( \rho^{-1} \lambda^{-1} \poly(\log(n/\delta))\right)
\end{align*}
iterations. 
Consider loss function $\ell:\mathbb{R}\times \mathbb{R}\rightarrow [0,1]$ that is 1-Lipschitz in its first argument.
Then with probability at least $1-\delta$ over the random initialization on $U(0)\in \mathbb{R}^{d\times m}$
and $a\in \mathbb{R}^m$ and the training samples,the population loss $L_{\mathcal{D}}(f):=\E_{(x,y)\sim \mathcal{D}}[\ell(f(U(T),x,a),y)]$ is upper bounded by
\begin{align*}
L_{\mathcal{D}}(f)\leq \sqrt{ {2y^\top (H^{\infty})^{-1}y} / {n}} + O (\sqrt{ \log ( n/ ( \lambda \delta ) )/ {(2n)}} ).
\end{align*}
\end{theorem}
\begin{proof}
We will define a sequence of failing events and bound these failure probability individually,
then we can apply the union bound to obtain the desired result.

Let $E_1$ be the event that $\lambda_{\min}(H^{\infty})<\lambda$.
Because $\mathcal{D}$ is $(\lambda,\delta/3,n)$-non-degenerate,
$\Pr[E_1]\leq \epsilon/3$.
In the remaining of the proof we assume $E_1$ does not happen.

Let $E_2$ be the event that $L_S(f(U(T),\cdot,a))=\frac{1}{n}\sum_{i=1}^n \ell(f(U(T),x_i,a),y_i)>\frac{1}{\sqrt{n}}$.
By Theorem \ref{thm:fl-ntk-convergence} with scaling $\delta$ properly, with probability $1-\delta/9$ we have $L_S(f(U(T),\cdot,a)) \leq \frac{1}{\sqrt{n}}$.
So we have $\Pr[E_2]\leq \delta/9$.

Set $R,B>0$ as
\begin{align*}
R= & ~O(\frac{n\sqrt{\log(m/\delta) \log^2(n/\delta)}}{\lambda\sqrt{m}}),\\
B= & ~(y^\top (H^{\infty})^{-1} y)^{1/2}+O\left( \frac{n\sigma }{\lambda}\cdot \poly(\log(m/\delta))+\frac{n^{7/2}}{\sigma^{1/2} {m}^{1/4}} \cdot \poly(\log(m/\delta))\right).
\end{align*}

Notice that $\|y\|_2=O(\sqrt{n})$ and $\|(H^{\infty})^{-1}\|_2=1/\lambda$.
By our setting of $\sigma=O(\frac{\lambda \poly(\log n,\log(1/\delta))}{n})$ and $m\sigma^2\geq n^{14}>n^{12}$,
 $B=O(\sqrt{n/\lambda})$.
Let $E_3$ be the event that there exists $r\in [m]$ so that $\|u_r-u_r(0)\|_2> R$, or $\|U-U(0)\|_F> B$.
By Lemma \ref{lem:total_movement}, $\Pr[E_3]\leq \delta/9$.

For $i=1,2,\cdots,$,
let $B_i=i$.
Let $E_4$ be the event that there exists $i>0$ so that 
\begin{align*}
\mathcal{R}_S(\mathcal{F}_{R,B_i}^{U(0),a})> \frac{B_i}{\sqrt{2n}}\left(1+(\frac{2\log(18/\delta)}{m})^{1/4}\right)+\frac{2R^2\sqrt{m}}{\sigma}+R\sqrt{2\log(18/\delta)}.
\end{align*}
By Lemma \ref{lem:rademacher_upper_bound},
$\Pr[E_4]\leq 1-\delta/9$.

Assume neither of $E_3,E_4$ happens.
Let $i^*$ be the smallest integer so that $B_{i^*}=i^*\geq B$,
then we have $B_{i^*}\leq B+1$ and $i^*=O(\sqrt{n/\lambda})$.
Since  $E_3$ does not happen,
we have $f(U(T),\cdot,a)\in \mathcal{F}_{R,B_{i^*}}^{U(0),a}$.
Moreover,
\begin{align*}
\mathcal{R}_S(\mathcal{F}_{R,B_{i^*}}^{U(0),a})
\leq & ~\frac{B+1}{\sqrt{2n}}\left(1+(\frac{2\log(18/\delta)}{m})^{1/4}\right)+\frac{2R^2\sqrt{m}}{\sigma}+R\sqrt{\log(18/\delta)}\\
= & ~\sqrt{\frac{y^\top (H^{\infty})^{-1}y}{2n}}+\frac{1}{\sqrt{n}}+O(\frac{\sqrt{n}\sigma\cdot \poly(\log(m/\delta))}{\lambda})\\
+ & ~\frac{n^{3}\poly(\log m,\log(1/\delta),\lambda^{-1})}{m^{1/4}\sigma^{1/2}}+\frac{2R^2\sqrt{m}}{\sigma}+R\sqrt{\log(18/\delta)}\\
= & ~\sqrt{\frac{y^\top (H^{\infty})^{-1}y}{2n}}+\frac{1}{\sqrt{n}}+O(\frac{\sqrt{n}\sigma\cdot \poly (\log (m/\delta))}{\lambda})+\frac{n^{3}\poly(\log m,\log(1/\delta),\lambda^{-1})}{m^{1/4}\sigma^{1/2}}\\
= & ~\sqrt{\frac{y^\top (H^{\infty})^{-1}y}{2n}}+\frac{2}{\sqrt{n}}
\end{align*}
where the first step follows from $E_4$ does not happen and the choice of $B$,
the second step follows from the choice of $R$,
and the last step follows from the choice of $m$ and $\sigma$.

Finally,
let $E_5$ be the event so that there exists $i\in\{1,2,\cdots,O(\sqrt{n/\lambda})\}$ so that
\begin{align*}
\sup_{f\in \mathcal{F}_{R,B_i}^{U(0),a}} \{L_{\mathcal{D}}(f)-L_S(f)\} > 2\mathcal{R}_S(\mathcal{F}_{R,B_i}^{U(0),a})+\Omega \left(\sqrt{\frac{\log( \frac{n}{\lambda \delta}) }{2n}}\right).
\end{align*}
By Theorem \ref{thm:sample_to_generalization} and applying union bound on $i$,
we have $\Pr[E_5]\leq \delta/3$.

In the case that all of the bad events $E_1,E_2,E_3,E_4,E_5$ do not happen,
\begin{align*}
L_{\mathcal{D}}(f(U(T),\cdot,a))
\leq & ~L_S(f(U(T),\cdot,a))+2\mathcal{R}_S(\mathcal{F}_{R,B_{i^*}}^{U(0),a})+O\left(\sqrt{\frac{\log ( \frac{n}{\lambda \delta} ) }{2n}}\right)\\
\leq & ~\sqrt{\frac{2y^\top (H^{\infty})^{-1}y}{n}}+\frac{5}{\sqrt{n}}+O\left(\sqrt{\frac{\log ( \frac{n}{\lambda \delta} ) }{2n}}\right)\\
= & ~\sqrt{\frac{2y^\top (H^{\infty})^{-1}y}{n}}+O\left(\sqrt{\frac{  \log ( \frac{n}{\lambda \delta} ) }{2n}}\right).
\end{align*}
which is exactly what we need.
\end{proof}

\newpage
\ifdefined\isarxiv
\bibliographystyle{alpha}%{alpha} %%%% Zhao : this line controls the bib style, we can use {plain} which is the 1,2,3, version
\else
\bibliographystyle{icml2021}
\fi
\bibliography{ref}

\end{document}